\documentclass[final,5p,times,twocolumn]{elsarticle}

\usepackage{amssymb}
\usepackage{amsmath}

\usepackage{cite}
\usepackage{amsmath,amssymb,amsfonts}
\usepackage{algorithmic}
\usepackage{graphicx}
\usepackage{textcomp}

\usepackage{booktabs}
\usepackage{multirow}
\usepackage{makecell}
\usepackage{amssymb}
\usepackage{amsmath}
\usepackage{array}
\usepackage{bm}
\usepackage{tabularx}
\usepackage{adjustbox}
\usepackage{caption}
\usepackage{subcaption}

\journal{Medical Image Analysis}

\begin{document}

\begin{frontmatter}

\title{Geometrical Cross-Attention and Nonvoid Voxelization for Efficient 3D Medical Image Segmentation}

\cortext[cor1]{Co-first authors.}
\cortext[cor2]{Corresponding author.}

\author[0,1]{Chenxin Yuan\corref{cor1}} \ead{chenxinyuan@ieee.org}
\author[0]{Shoupeng Chen\corref{cor1}} \ead{spchen1207@std.uestc.edu.cn}
\author[2]{Haojiang Ye} \ead{haojiangye@link.cuhk.edu.cn}
\author[0]{Yiming Miao\corref{cor2}} \ead{yimingmiao@ieee.org}
\author[0,3]{Limei Peng} \ead{auroraplm@knu.ac.kr}
\author[0,4]{Pin-Han Ho} \ead{p4ho@uwaterloo.ca}

\affiliation[0]{organization={Shenzhen Institute for Advanced Study},
            addressline={University of Electronic Science and Technology of China}, 
            city={Shenzhen},
            postcode={518100}, 
            state={Guangdong},
            country={China}}

\affiliation[1]{organization={School of Electrical and Electronic Engineering},
            addressline={Nanyang Technological University}, 
            city={Singapore},
            postcode={639798}, 
            country={Singapore}}

\affiliation[2]{organization={School of Science and Engineering},
            addressline={The Chinese University of Hong Kong, Shenzhen}, 
            city={Shenzhen},
            postcode={518172}, 
            state={Guangdong},
            country={China}}

\affiliation[3]{organization={School of Computer Science and Engineering},
            addressline={Kyungpook National University}, 
            city={Daegu},
            postcode={37224}, 
            country={South Korea}}
            
\affiliation[4]{organization={Department of Electrical and Computer Engineering},
            addressline={University of Waterloo},
            city={Waterloo},
            postcode={N2L3G1}, 
            state={Ontario},
            country={Canada}}

\begin{abstract}
Accurate segmentation of 3D medical scans is crucial for clinical diagnostics and treatment planning, yet existing methods often fail to achieve both high accuracy and computational efficiency across diverse anatomies and imaging modalities. To address these challenges, we propose GCNV-Net, a novel 3D medical segmentation framework that integrates a Tri-directional Dynamic Nonvoid Voxel Transformer (3DNVT), a Geometrical Cross-Attention module (GCA), and Nonvoid Voxelization. The 3DNVT dynamically partitions relevant voxels along the three orthogonal anatomical planes, namely the transverse, sagittal, and coronal planes, enabling effective modeling of complex 3D spatial dependencies. The GCA mechanism explicitly incorporates geometric positional information during multi-scale feature fusion, significantly enhancing fine-grained anatomical segmentation accuracy. Meanwhile, Nonvoid Voxelization processes only informative regions, greatly reducing redundant computation without compromising segmentation quality, and achieves a 56.13\% reduction in FLOPs and a 68.49\% reduction in inference latency compared to conventional voxelization. We evaluate GCNV-Net on multiple widely used benchmarks: BraTS2021, ACDC, MSD Prostate, MSD Pancreas, and AMOS2022. Our method achieves state-of-the-art segmentation performance across all datasets, outperforming the best existing methods by 0.65\% on Dice, 0.63\% on IoU, 1\% on NSD, and relatively 14.5\% on HD95. All results demonstrate that GCNV-Net effectively balances accuracy and efficiency, and its robustness across diverse organs, disease conditions, and imaging modalities highlights strong potential for clinical deployment.
\end{abstract}


\end{frontmatter}

\section{Introduction}
Segmentation of anatomical structures accurately from three-dimensional (3D) medical images is critical for clinical practice, assisting clinicians in diagnosis, surgical planning, targeted radiotherapy, and monitoring disease progression \citep{BraTS}. 
Despite its importance, manual segmentation is labor-intensive, time-consuming, and subject to inter-observer variability, underscoring the need for robust automated methods.

Recent advances in deep learning have greatly improved automated segmentation, often achieving performance comparable or superior to that of well-trained radiologists. 
Among these approaches, the U-Net architecture \citep{unet} has been particularly influential. Its symmetric encoder-decoder structure with multi-scale skip connections allows effective segmentation even with limited data. Variations of U-Net, through enabling 3D segmentation \citep{3dunet} and including attention mechanisms \citep{attenUnet}, have further enhanced segmentation quality and modality. The nnU-Net framework \citep{nnUnet}, which adapts preprocessing and network configurations automatically to new datasets, currently stands as a competitive baseline for medical segmentation tasks.

More recently, Transformer-based models, which utilize self-attention mechanisms to capture global relationships, have shown potential in achieving even higher segmentation accuracy. Approaches such as UNETR \citep{unetr} introduced a purely Transformer-based encoder, while SwinUNETR \citep{swinUnetr} integrate hierarchical Swin Transformers into medical segmentation architectures. Hybrid approaches like TransBTS \citep{transBTS} and nnFormer~\citep{nnformer} combine CNNs and Transformers, effectively leveraging both local details and global relationships. 

However, current methods face two key limitations. First, while CNNs like nnU-Net can effectively capture spatial context through large receptive fields, this spatial awareness is implicit. It emerges from stacked convolutions rather than being explicitly encoded at each processing stage. Similarly, positional encoding schemes commonly used in Transformers (sinusoidal~\citep{vit}, RoPE~\citep{rope}) inject position information at the input stage, but this information is not explicitly maintained as features pass through multi-scale fusion and down/up-sampling operations. Second, standard 3D architectures process every voxel uniformly, expending identical computation on background and anatomically relevant regions alike, which limits deployment in resource-constrained clinical settings. These two limitations motivate a framework that improves both geometric awareness and computational efficiency simultaneously.

To overcome this limitation, we propose a novel Geometrical Cross-Attention (GCA) module, explicitly incorporating geometric positional context during feature fusion. This approach yields substantial improvements in segmentation accuracy with gains of 1.88\% (Dice) and 25.3\% (HD95), especially in detail-rich and anatomically complex regions.

Meanwhile, to achieve computational savings without compromising accuracy, we introduce a Nonvoid Voxelization strategy that selectively allocates computation to anatomically relevant regions. This strategy delivers substantial efficiency gains essentially ‘for free,’ enabling faster segmentation without loss of accuracy. In parallel, we develop a Tri-directional Dynamic Nonvoid Voxel Transformer (3DNVT), which reduces the computational time complexity of Transformer-based 3D medical image segmentation from sextic to quartic.

In this work, we present GCNV-Net, a 3D medical image segmentation framework that simultaneously addresses both accuracy and efficiency challenges through three key innovations: a Nonvoid Voxelization strategy, the 3DNVT architecture, and GCA modules. These components are designed to effectively capture complex spatial dependencies within irregularly structured medical voxels. Collectively, they enable GCNV-Net to achieve superior segmentation accuracy with substantial computational efficiency, making it well suited for practical clinical applications. The main contributions of this work are summarized as follows:

\begin{itemize}
    \item We introduce a Nonvoid Voxelization strategy and regularization that substantially reduces computational redundancy by selectively processing only anatomically relevant regions within 3D medical image volumes. Ablation studies demonstrate that, compared with traditional voxelization, this technique achieves a reduction of 47.31\%-85.76\% in embedded voxels, 56.31\% fewer FLOPs, and 68.49\% shorter inference latency, all without compromising segmentation accuracy under the same model capacity.
    \item We propose a Tri-directional Dynamic Nonvoid Voxel Transformer (3DNVT), specifically designed to efficiently model spatial context by dynamically partitioning sparse, anatomically informative voxels along three orthogonal planes -- transverse, sagittal, and coronal. This tailored Transformer effectively captures complex spatial dependencies, yielding a 2.43\% (Dice) and 29.1\% (HD95) improvement in segmentation accuracy over using only conventional Transformers.
    \item We develop a Geometrical Cross-Attention (GCA) module that explicitly incorporates geometric positional information during multi-scale feature fusion. By enhancing spatial representation learning, the GCA module improves segmentation accuracy by 1.88\% (Dice) and 25.3\% (HD95), particularly in detail-rich and anatomically complex regions.
\end{itemize}

\section{Related Work}
\subsection{Efficient Segmentation Architectures}
Beyond improving segmentation quality, balancing the computational efficiency of segmentation models remains an important practical consideration. Previous efforts to improve efficiency have explored lightweight architectures and sparse computation strategies. 
Methods like CoTr \citep{cotr} and MedNeXt \citep{mednext} reduce computations via feature downsampling, while approaches such as 3D UX-Net \citep{3duxnet} and SegFormer3D \citep{SegFormer3D} utilize depth-wise convolutions and lightweight hierarchical Transformers. E2ENet \citep{e2enet} and SDV-TUNet \citep{SDVTUNet} achieve the goal through sparse processing, and LHU-Net \citep{lhunet} combines spatial and channel attention at different network depths to achieve competitive accuracy with substantially fewer parameters. However, although these methods effectively reduce FLOPs, they often compromise segmentation quality and fail to strike a balance between accuracy and efficiency, performing particularly poorly in anatomically complex regions, which limits their clinical applicability.

\subsection{Spatial and Geometric Context Modeling}
Capturing and keeping spatial context and geometric information is central to delineating fine anatomical boundaries in 3D segmentation. For this purpose, transformer architectures commonly introduce positional information via sinusoidal encodings \citep{vit} with extensions such as rotary position embeddings (RoPE) that couple absolute positions with relative phase rotation \citep{rope}. Although effective for image perception, these schemes are largely agnostic to the explicit Euclidean organization of volumetric anatomy and usually contribute only strongly at early layers. Their influence often degrades rapidly during feature fusion, weakening geometric consistency. In fact, evidence from other 3D perception domains suggests that explicitly coupling attention mechanisms with geometric information can enhance boundary localization. For example, sparse key-point sampling in Deformable DETR \citep{DeformableDERT} for object detection and relative offset injection in Point Transformer \citep{pointTransformer} for point cloud analysis both improve spatial localization accuracy. However, in medical image, existing deep learning-based methods still fail to explicitly associate extracted features with their corresponding geometric positions, leaving this as an open challenge that demands further investigation.

\begin{figure*}[ht]
    \centering
    \includegraphics[width=\textwidth]{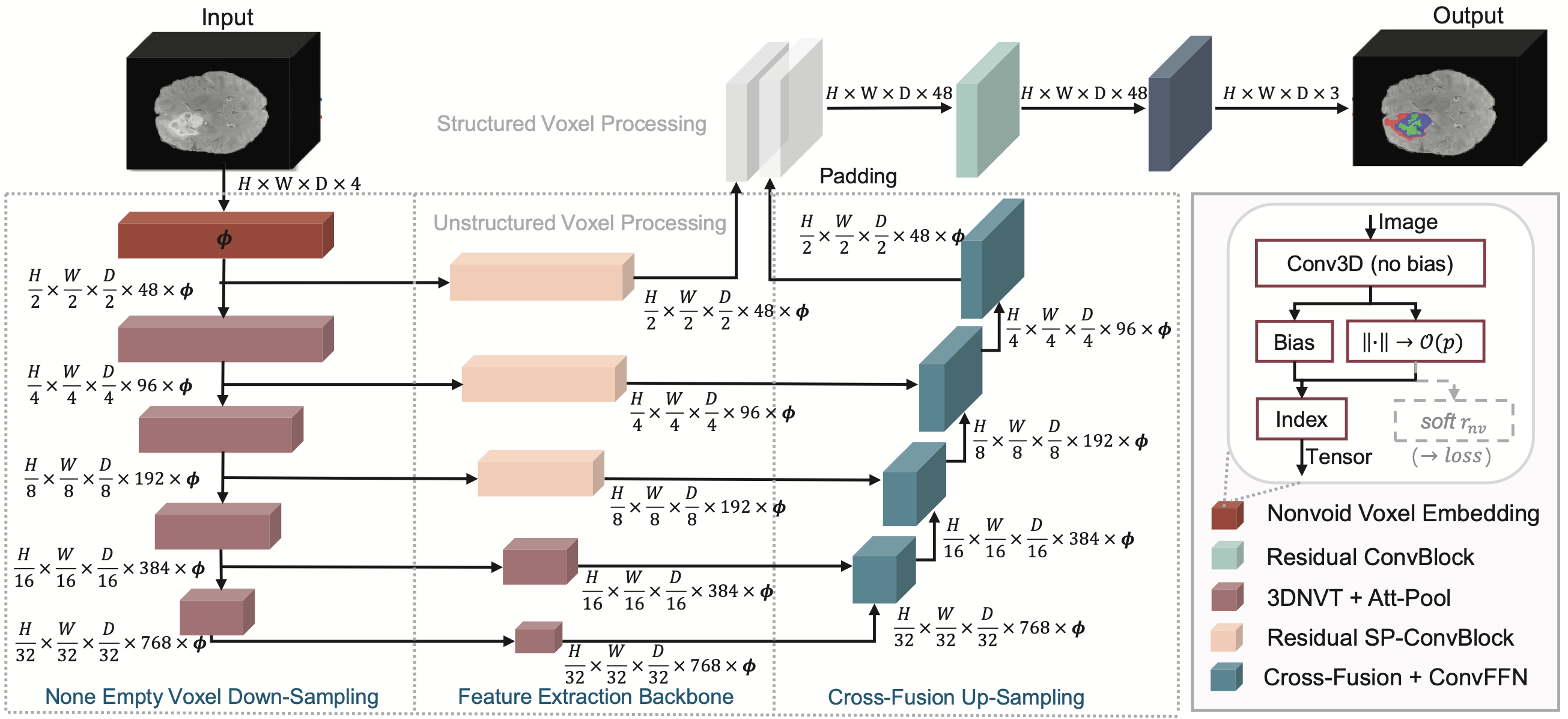}
    \caption{Overview of the proposed GCNV-Net. The Nonvoid Voxelization converts dense volumes to sparse nonvoid voxels. The encoder applies 3DNVT blocks with GCA-based downsampling. The decoder fuses multi-scale features using GCA-based upsampling and residual convolution to produce dense segmentation.}
    \label{Metho_Fig_overallArchi}
\end{figure*}
\subsection{Sparse 3D Processing and Dynamic Window Partition}
Transformer encoders have emerged as strong backbones for volumetric segmentation. However, straightforward 3D Transformers are computationally prohibitive, since attention within a $t \times t \times t$ window scales with a time complexity of $O(t^6)$. To address this, much of the prior work in 3D perception has focused on improving efficiency through sparsity or structured locality.
Sparse convolution methods, such as submanifold sparse convolutions \citep{submanifold} and Minkowski engine \citep{Minkowski}, limit computation to active sites, while large-scale LiDAR approaches like SPVNAS \citep{spvnas} further leverage the inherent sparsity of point clouds. Building on these ideas, DSVT \citep{dsvt} introduces density-aware token grouping with adaptive windowing for point-cloud detection, and OneFormer3D \citep{Oneformer3d} unifies multiple 3D recognition tasks with a Transformer that integrates both local and global pathways over sparse inputs.
Despite these advances, medical volumes pose a distinct challenge. Unlike outdoor or indoor scenes where sparsity is relatively uniform, 3D medical scans exhibit dense and irregular clusters of anatomically meaningful voxels embedded within vast, uninformative backgrounds. As a result, sparse processing strategies developed for uniformly sparse data -- such as those used in point cloud analysis -- cannot be directly applied to medical image. Bridging this fundamental mismatch remains an open research problem.

\section{Methodology}

The overall architecture of the proposed GCNV-Net is depicted in Fig.~\ref{Metho_Fig_overallArchi}, which follows a typical U-Net framework and designed specifically for 3D medical segmentation. It incorporates sparse nonvoid voxel processing and explicit geometrical context attention to improve both segmentation accuracy and computational efficiency.

\begin{figure}
    \centering
    \includegraphics[width=3.1in]{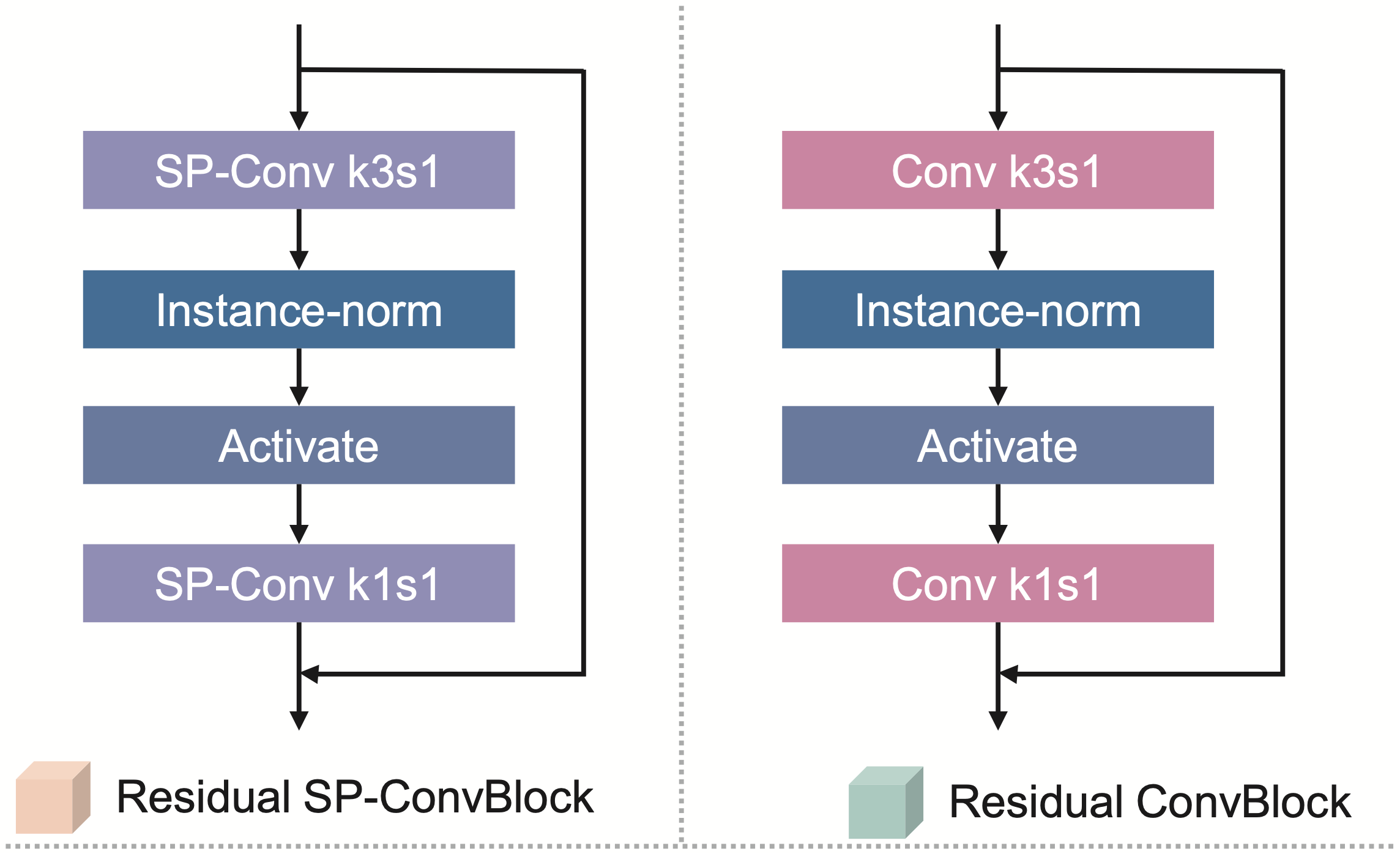}
    \caption{Feature extraction backbones. (a) Residual SP-ConvBlock: sparse residual convolution blocks on sparse nonvoid voxels. (b) Residual ConvBlock: conventional residual convolution blocks on structured dense volumes.}
    \label{Metho_Fig_resConvBlock}
\end{figure}

The process begins with a Nonvoid Voxelization module, which extracts only anatomically informative voxels from the input 3D medical volumes. These nonvoid voxels are then processed through a hierarchical encoder consisting of four stages, each comprising a 3DNVT block followed by a GCA downsampling module, aiming to effectively reduce spatial resolution while retaining structural and spatial integrity. 
After each GCA downsampling, sparse features at multiple scales subsequently undergo some sparse processing for further feature extraction. These blocks operate directly on sparse voxel representations, preserving geometric information and enabling efficient computation. Specifically, considering latency and memory consumption, features at high resolution will be passed through some residual convolutional blocks, as shown in Fig.~\ref{Metho_Fig_resConvBlock}. Features at low resolution, on the other hand, will be processed by some 3DNVT blocks.
Finally, in the decoder path, multi-scale features are fused via a GCA upsampling module. A final convolutional block is then used to produce the segmentation output.

\subsection{Nonvoid Voxelization}
In volumetric medical imaging, a substantial fraction of the input volume is occupied by uninformative background, such as air surrounding the body in thoracic CT, or zero-intensity regions outside the skull in brain MRI. Standard 3D architectures process every voxel uniformly, expending identical computation on background and anatomically relevant regions alike. Although prior works have introduced attention mechanisms to guide the network's focus toward possibly more informative regions, these approaches merely down-weight irrelevant voxels rather than eliminating them from the computational graph, where every voxel still incurs the full cost of forward propagation. To move beyond soft reweighting, we introduce Nonvoid Voxelization (NV), a simple mechanism that explicitly identifies and discards void voxels at the earliest network stage, enabling all downstream layers to operate exclusively on clinically meaningful content.

\textbf{Bias-free Embedding and Nonvoid Detection.}
As illustrated in the inset of Fig.~\ref{Metho_Fig_overallArchi}, given an input volume $\mathcal{X} \in \mathbb{R}^{H \times W \times D \times M}$, where $H$, $W$, $D$, and $M$ denote the height, width, depth, and number of modalities, respectively, we first embed the input into voxel features through a 3D convolution with kernel size $k$, stride $s$, output channels $C$, and without bias:
\begin{equation}
    \mathcal{F} = \mathrm{Conv3D}_{k,s,C}(\mathcal{X} - b),
    \label{eq:nve_conv}
\end{equation}
where $\mathcal{F} \in \mathbb{R}^{H' \times W' \times D' \times C}$ is the resulting feature map, $H'$, $W'$, $D'$ are the spatial dimensions after convolution, and $b \in \mathbb{R}^{M}$ is a per-channel background constant derived from the normalization statistics of the preprocessing pipeline (see Supplementary~Material~Section~4.1 for details on modality-specific derivations). When the normalization naturally preserves zero-valued background, $b=0$ and Eq.~\eqref{eq:nve_conv} reduces to a plain bias-free convolution.

Note that the absence of bias in the embedding convolution is essential. Consider a spatial patch whose input voxels are all equal to the background value $b$: after centering, $\mathcal{X} - b$ will become identically zero, and a bias-free convolution maps a zero input to a strictly zero output. This deterministic zero-output property is what allows nonvoid detection to work, where background and foreground can be reliably separated by simply checking whether the output feature magnitude exceeds a small threshold.

Following embedding, we compute an occupancy map $\mathcal{O} \in \{0,1\}^{H' \times W' \times D'}$ based on the $\ell_p$ norm of each feature vector:
\begin{equation}
  \mathcal{O}[x, y, z] =
  \begin{cases}
    1, & \text{if } \|\mathcal{F}[x, y, z, :]\|_p > \epsilon, \\
    0, & \text{otherwise},
  \end{cases}
  \label{eq:occupancy}
\end{equation}
where $c = (x, y, z) \in \mathbb{R}^3$ represents the geometric coordinates of each voxel, and $\epsilon$ is a small positive threshold (e.g., $10^{-5}$ in our experiments). Since the zero-output property follows directly from the linearity of bias-free convolution (a zero input always produces a zero output when no bias is present), the hyperparameter $\epsilon$ here only serves as a numerical guard against floating-point rounding rather than a sensitive hyperparameter (a sensitivity analysis can be found in Supplementary~Material~Section~5).

Then, from the occupancy map $\mathcal{O}$, we extract the coordinates $\mathcal{C}$ of all nonvoid voxels:
\begin{equation}
  \mathcal{C} = \{(x_i, y_i, z_i) \mid
  \mathcal{O}[x_i, y_i, z_i] = 1\}_{i=1}^{\phi},
\end{equation}
where $\phi$ denotes the number of nonvoid voxels, which varies across input volumes. 

Finally, these spatial locations are then used to gather the corresponding feature vectors $f_i \in \mathbb{R}^{C}$ from the feature map $\mathcal{F}$:
\begin{equation}
  f_i = \mathcal{F}[x_i, y_i, z_i, :],
\end{equation}
yielding a set of sparse nonvoid voxel embeddings:
\begin{equation}
  \mathcal{V} = \{v_i \mid v_i = (\mathbf{c}_i, f_i, d_i)\}_{i=1}^{\phi},
\end{equation}
where $d_i$ is a unique voxel index. 

Unlike conventional Transformer-based segmentation methods that rely solely on relative sinusoidal positional encodings, this explicit retention of absolute geometric coordinates preserves accurate positional context throughout entire downstream sparse processing stages.

\textbf{Soft Nonvoid Regularization.}
The hard thresholding in Eq.~\eqref{eq:occupancy} is non-differentiable and cannot provide gradient signal to the embedding convolution. To encourage the network to actively suppress feature energy in background regions, we introduce a soft nonvoid ratio $r_{\mathrm{nv}}$ as an auxiliary regularization term. Specifically, $r_{\mathrm{nv}}$ is computed as a differentiable approximation of the fraction of voxels classified as nonvoid:
\begin{equation}
   r_{\mathrm{nv}} = \frac{1}{N} \sum_{i=1}^{N}
   \sigma\!\left(\frac{\|\mathcal{F}(i)\|_p - \epsilon}{\tau}\right),
   \label{eq:soft_nv}
\end{equation}
where $N$ is the total number of embedded voxels, $\sigma(\cdot)$ is the sigmoid function, and $\tau$ is a temperature that controls the sharpness of the soft gating. As $\tau \to 0$, Eq.~\eqref{eq:soft_nv} recovers the hard occupancy map in Eq.~\eqref{eq:occupancy}. 

The overall training cost is then:
\begin{equation}
   \mathcal{L}_{\mathrm{total}} = \mathcal{L}_{\mathrm{seg}} +
   \lambda_{\mathrm{nv}} \, r_{\mathrm{nv}},
   \label{eq:total_loss}
\end{equation}
where $\mathcal{L}_{\mathrm{seg}}$ is the standard segmentation loss (Dice + Cross-Entropy) and $\lambda_{\mathrm{nv}}$ is a hyperparameter controlling the regularization strength. Minimizing $r_{\mathrm{nv}}$ provides continuous optimization pressure, enabling the network to produce near-zero responses for background while preserving strong activations in task-relevant regions. During inference, only the hard occupancy mask $\mathcal{O}$ is applied, so the soft term introduces no additional cost at test time.

\subsection{Tri-directional Dynamic Nonvoid Voxel Transformer}
\begin{figure}
    \centering
    \includegraphics[width=3in]{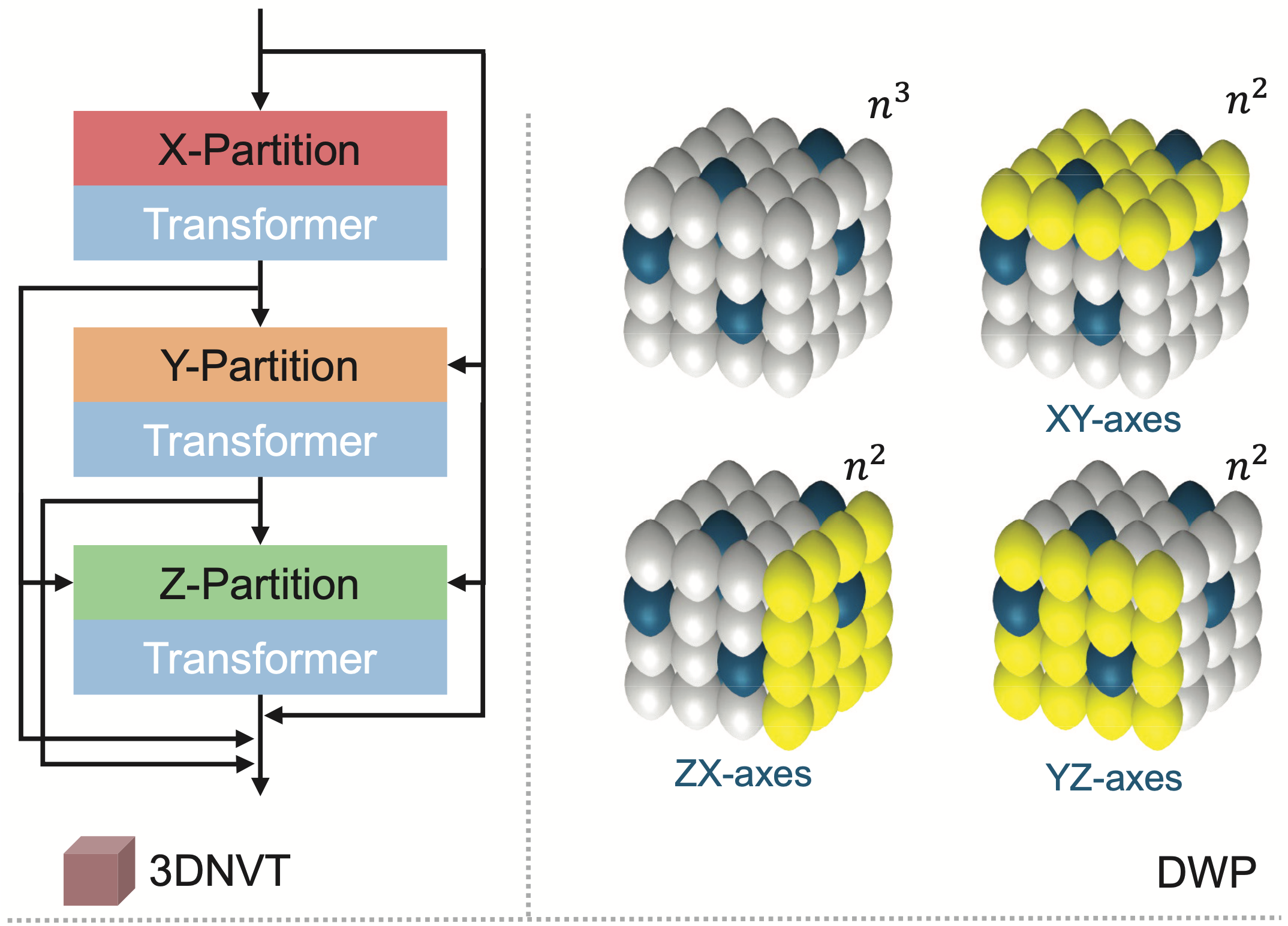}
    \caption{Tri-directional Dynamic Nonvoid Voxel Transformer (3DNVT): dynamic window partitioning followed by plane-wise (transverse/sagittal/coronal) voxel grouping}
    \label{Metho_Fig_3DNVT}
\end{figure}

Transformer-based architectures have recently demonstrated remarkable performance in multimodal 3D medical image segmentation \citep{swinUnetr, unetr}. However, the computational time complexity of traditional 3D Transformers grows drastically with window size, following a sextic order $(O(t^6))$ in terms of window width $t$, significantly limiting the feasibility for practical clinical use. Although the proposed Nonvoid Voxelization dramatically reduces computation without degrading segmentation accuracy, due to the native characteristic of 3D medical images, these nonvoid voxels typically form densely packed, irregular clusters rather than being uniformly sparse. As a result, existing sparse processing approaches tailored to uniformly sparse data, such as outdoor point clouds \citep{Oneformer3d, dsvt}, cannot be directly applied. To bridge this methodological gap, we introduce a novel Tri-directional Dynamic Nonvoid Voxel Transformer block, as illustrated in Fig.~\ref{Metho_Fig_3DNVT}.

Specifically, given a set of nonvoid voxels $\mathcal{V}$ partitioned into non-overlapping 3D windows, each window $j$ comprises a subset of voxels defined as:
\begin{equation}
    \mathcal{V}_j = \{v_i \mid v_i=(c_i,f_i,d_i)\}_{i=1}^{\phi_j}.
\end{equation}

Unlike fixed-size window processing that inefficiently handles densely clustered voxels, we dynamically partition each 3D window into smaller subsets, 
where the number of subsets $\mathcal{S}$ adapts to voxel density, calculated as:
\begin{equation}
    \mathcal{S}_j = \lfloor \frac{\phi_j}{\tau} \rfloor + \mathbb{I}[\phi_j\mod\tau > 0],
\end{equation}
where $\tau$ denotes the maximum number of voxels per subset, and $\mathbb{I}[\cdot]$ is the indicator function. 

To capture direction-specific spatial dependencies, we subsequently partition these subsets along the three standard orthogonal anatomical planes: transverse plane (XY), sagittal plane (YZ), and coronal plane (XZ). This is done by sorting the voxels along the corresponding axis. For instance, when partitioning along the Z-axis yields, the $s$-th subset of the XY-plane is:
\begin{equation}
    \mathcal{Z}_{j}^{s} = \{ v_i \mid d_{z,i} \in [(s-1)\tau, s\tau) \}_{i=1}^{\tau -1},
\end{equation}
where $s \in 1,2,\dots,\mathcal{S}_j$ and $d_{z,i}$ is the temporary index of voxel $v_i$ after sorted along the Z-axis. Equivalent partitions are applied to the XZ- and YZ-planes by sorting along the Y- and X-axes, respectively.

By introducing such an axis-specific subdivision, we finally reduce the sequence length of each subset from $O(n^3)$ to $O(n^2)$, hence decreasing time complexity of Transformer-based 3D medical segmentation from $O(n^6)$ to $O(n^4)$. Moreover, the dynamic window partition allows adaptive resource allocation where densely populated windows will dynamically produce more subsets and be allocated more computational resources, resulting in more accurate segmentation.

\subsection{Geometrical Cross-Attention}
\begin{figure}[t]
    \centering
    \includegraphics[width=3.3in]{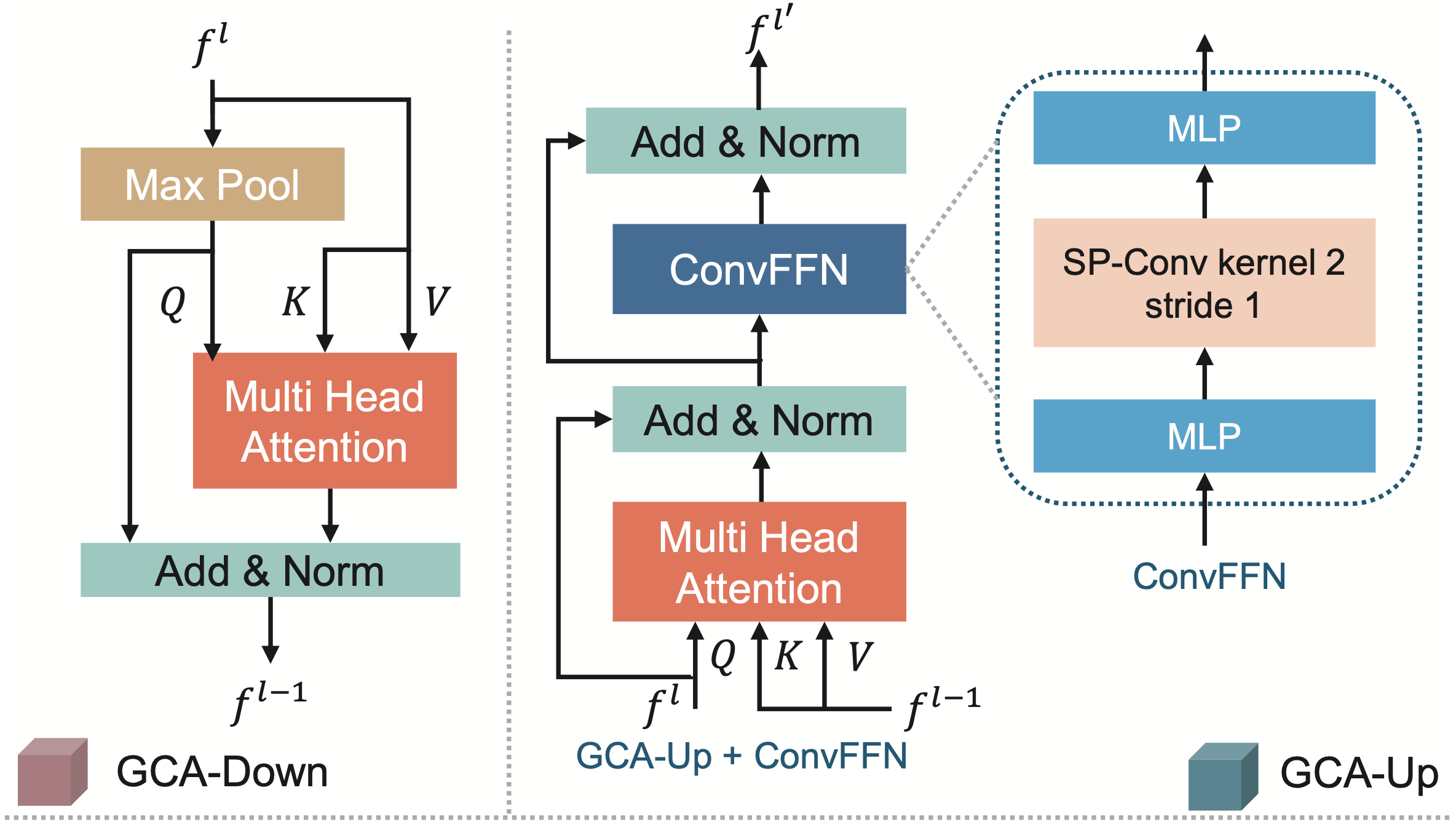}
    \caption{Geometrical Cross-Attention (GCA) modules. GCA-Down uses pooled features as queries with high-resolution features as keys/values, and GCA-Up reverses the direction for decoder fusion.}
    \label{Metho_Fig_GeoAtten}
\end{figure}

Multi-scale feature fusion is a core operation of the U-Net architecture. However, standard max pooling selects the maximum value within each window but discards the spatial correspondence between the selected value and its original position. Similarly, trilinear upsampling interpolates features without reference to the geometric positions of the sparse voxels being reconstructed. As a result, positional context injected at early stages is progressively lost across resolution changes, degrading fine-grained segmentation accuracy \citep{DeformableDERT,pointTransformer}. To address this, we propose a novel Geometrical Cross-Attention mechanism, specifically designed for effective feature fusion across different scales. It explicitly encodes geometric information during both downsampling and upsampling. As given in Fig.~\ref{Metho_Fig_GeoAtten}, the proposed mechanism is included in two key modules: GCA-Down for downsampling and GCA-Up for upsampling.

\textbf{GCA-Down.} During downsampling, nonvoid voxels are inherently unstructured and irregularly distributed. Given a cubic window $w$ containing $P$ nonvoid voxels, we first apply standard max pooling to generate coarse feature representations:

\begin{equation}
    f_{w,coarse}^{l-1} = MaxPool(\{f_{w,i}^l\}^{{P}}_{i=1}),
\end{equation}
where $f_{w,coarse}^{l-1} \in \mathbb{R}^{C}$ is the pooled feature at the lower layer (i.e., layer $l-1$)  with lower resolution.

However, pooling alone lacks spatial awareness. To preserve geometry relationships, we apply cross-attention using pooled coarse features as queries (Q) and high-resolution voxel features (with their explicit coordinates embedded) as keys (K) and values (V):
\begin{equation}
\begin{split}
    f_w^{l-1} = &\text{MHA}(\text{Q}=f_{w,coarse}^{l-1}+\text{PE}(c^{l-1}), \\
              &\text{KV}=\text{MLP}(\{f_{w,i}^l+\text{PE}(c_i^{l})\}^{P}_{i=1})),
\end{split}
\end{equation}
where $\text{MHA}(\cdot)$ is the standard multi-head attention, $\text{PE}(\cdot)$ is the positional embedding, and $\text{MLP}(\cdot)$ is the multilayer perceptron for dimension and resolution alignment.

Finally, we have the downsampled sparse voxel set:
\begin{equation}
    \mathcal{V}^{l-1}=\{v_i^{l-1} \mid v_i^{l-1}=(c_i^{l-1}, f_i^{l-1},d_i)\}_{i=1}^{\phi}.
\end{equation}

\textbf{GCA-Up.} During decoding, we reverse the process. Residual features from the encoder serve as queries, and coarse decoder features are the keys and values:
\begin{equation}
\begin{split}  
    {f_w^{l}}' = \text{MLP}(\text{MHA}(&\text{Q}=\text{MLP}(\{f_{w,i}^l+\text{PE}(c_i^{l})\}^{P}_{i=1})), \\
                         &\text{KV}=f^{l-1}_w+\text{PE}(c^{l-1}))).
\end{split}
\end{equation}
Hence, the final updated voxel after decoding becomes:
\begin{equation}
    {\mathcal{V}^{l}}'=\{{v_i^l}' \mid {v_i^l}'=(c_i^{l}, {f_i^{l}}', d_i)\}_{i=1}^{N}.
\end{equation}
In contrast to previous multi-scale fusion strategies relying primarily on early relative position encoding, our GCA modules preserve explicit spatial context among all steps. By aligning features across different scales based on both content and geometry information, our method has potential to produce more accurate segmentation boundaries, particularly in spatially complex or fine-detailed regions.

\section{Experiment}
In this section, we evaluate GCNV-Net from two perspectives. We first analyze the Nonvoid Voxelization module in isolation — measuring how many voxels it eliminates, whether the resulting sparse representation faithfully  preserves anatomical content, and whether the learned embedding transfers across datasets without retraining (Section~\ref{sec:nv_analysis}). We then assess the end-to-end segmentation performance of GCNV-Net against state-of-the-art methods on five benchmark datasets, covering both accuracy and computational efficiency (Sections~\ref{sec:seg_quality} \& \ref{sec:qua_eff_tradeoff}). All experiments are conducted on a Linux server equipped with an NVIDIA RTX 4090 GPU. Training details are provided in Supplementary~Material~Section~4.

\subsection{Nonvoid Analysis across Datasets} \label{sec:nv_analysis}
\begin{table*}[ht]
  \fontsize{9bp}{8bp}\selectfont 
  \centering
  \caption{Nonvoid Voxelization efficiency across eleven 3D medical image datasets. Non-zero Ratio: percentage of voxels with non-zero intensity. Cropped Ratio: percentage of voxels remaining after cropping to the minimum bounding box of all non-zero voxels. Nonvoid/Traditional Voxels (k): average embedded voxel count using our Nonvoid Voxelization or classical voxelization, respectively. Embedded Voxel Saving: 1 - Nonvoid Voxels / Traditional Voxels. $\uparrow$/$\downarrow$: preferred direction.}
  \label{Exp_Table_NonvoidStatistics}
  \vspace{1mm}
  \begin{tabular}{cccccccc}
    \toprule
    \makecell{\textbf{Dataset}}

      & \makecell{\textbf{Non-zero}\\\textbf{Ratio (\%) $\downarrow$}}
      & \makecell{\textbf{Cropped}\\\textbf{Ratio (\%) $\downarrow$}}
      & \makecell{\textbf{Nonvoid}\\\textbf{Voxels (k) $\downarrow$}}
      & \makecell{\textbf{Traditional}\\\textbf{Voxels (k) $\downarrow$}}
      & \makecell{\textbf{Embedded Voxel}\\\textbf{Saving (\%) $\uparrow$}} \\
    \midrule
    MSD Prostate   & 99.50 & 99.82 &  57.19 & 131.1 & 56.38 \\
    BraTS2021      & 16.54 & 63.18 &  38.84 & 262.1 & 85.18 \\
    ACDC           & 97.72 & 99.92 &   6.71 &  16.4	& 59.11 \\
    SIMON BIDS     & 79.78 & 97.62 & 138.10 & 262.1	& 47.31 \\
    SPIDER         & 52.20 & 47.27 & 110.53 & 262.1	& 57.83 \\
    ISLES2022      & 19.10 & 70.77 &   2.58	&  16.4	& 84.26 \\
    MSD BrainTumor & 16.11 & 64.73 &  37.32 & 262.1	& 85.76 \\
    IXI            & 67.73 & 73.62 &  97.00 & 262.1	& 62.99 \\
    BreastDM       & 49.98 & 57.96 &  14.28 &  32.8	& 56.47 \\
    MSD Pancreas   & 48.75 & 86.77 & 118.18 & 524.3	& 77.46 \\
    AMOS2022       & 40.40 & 46.81 & 468.72 & 2097.2 & 77.65 \\
    \bottomrule
  \end{tabular}
\end{table*}

A fundamental yet often overlooked source of inefficiency in 3D medical image segmentation is the processing of uninformative background voxels. As we know, the computation load of deep nets usually rise linearly or quadratically with the size of the input voxels. Decreasing number of input voxels directly result in computation reduction. Hence, before evaluating GCNV-Net as a whole, we first quantify the extent of this redundancy across a broad range of datasets. This analysis serves two purposes: it motivates Nonvoid Voxelization as a general strategy for any volumetric segmentation pipeline, and it characterizes the efficiency ceiling that our Nonvoid Voxelization module can achieve in practice. Results are summarized in Table~\ref{Exp_Table_NonvoidStatistics}, our Nonvoid Voxelization can significantly reduce redundant voxel processing, particularly in datasets with extensive anatomical-irrelevant regions.

\textbf{Nonvoid Region Distribution.} First, we characterize the anatomical information density across eleven commonly used 3D medical image datasets, including MSD Prostate (prostate) \citep{MSD_Prostate}, BraTS2021 (brain) \citep{BraTS2021}, ACDC (cardiac) \citep{ACDC}, SIMON BIDS (brain) \citep{SIMON_BIDS}, SPIDER (spine) \citep{SPIDER}, ISLES2022 (brain) \citep{ISLES2022}, MSD BrainTumor (brain) \citep{MSD_BrainTumor}, IXI (brain) \citep{IXI}, BreastDM (chest) \citep{BreastDM}, MSD Pancreas (pancreas) \citep{MSD_Prostate}, and AMOS2022 (abdominal) \citep{AMOS2022}. For each dataset, we compute the non-zero ratio, defined as the number of voxels with non-zero intensity divided by the total number of voxels in the volume. As shown in Table~\ref{Exp_Table_NonvoidStatistics}, datasets such as BraTS2021 (16.54\%), ISLES2022 (19.10\%), MSD BrainTumor (16.11\%), AMOS2022 (40.40\%), MSD Pancreas (48.75\%), and BreastDM (49.98\%) exhibit low non-zero ratios (below 50\%), indicating that over half of each volume consists of zero-intensity background, which any standard segmentation network would process at full cost without contributing to the output. In contrast, MSD Prostate (99.50\%), ACDC (97.72\%), and SIMON BIDS (79.78\%) have typically high non-zero ratios, leaving relatively little zero-intensity background available for voxel-level pruning.

We additionally report the cropped ratio, defined as the number of voxels remaining after applying nnU-Net's standard foreground crop \citep{nnUnet} (i.e., cropping to the minimum bounding box of all non-zero voxels) divided by the total number of voxels. A high cropped ratio, such as ACDC (99.92\%) and MSD Prostate (99.82\%), indicates that nearly all voxels survive cropping, leaving little room for further reduction through boundary removal. Conversely, a low cropped ratio, as seen in AMOS2022 (46.81\%) and SPIDER (47.27\%), reveals that cropping does remove substantial border regions, hence, can meaningfully accelerate training. Generally,  cropping is a meaningful and effective strategy for efficiency, yet it has an inherent limitation: since it retains the full minimum bounding box, zero-valued regions outside the body but inside the bounding box (e.g., air surrounding the body) are preserved entirely. Furthermore, internal void regions within the body (e.g., air cavities or ventricles) are similarly untouched. On the other hand, voxel-level pruning, such as our Nonvoid Voxelization, operates at a finer granularity and can eliminate both types of residual void.

\textbf{Voxel Count Reduction.} Next, we compare the voxel counts produced by our Nonvoid Voxelization against those of traditional dense embedding (as used in~\citep{swinUnetr, SegFormer3D}). The embedded voxel saving is computed as $1 - N_{\text{nonvoid}} / N_{\text{traditional}}$, where $N_{\text{nonvoid}}$ and $N_{\text{traditional}}$ denote the average number of voxels produced by our method and by classical voxelization, respectively. As shown in Table~\ref{Exp_Table_NonvoidStatistics}, Nonvoid Voxelization achieves consistent and significant reductions across all datasets. In datasets with low non-zero ratios, such as BraTS2021 (16.54\% nonzero pixels, achieving 85.18\% voxel saving) and MSD BrainTumor (16.11\% nonzero pixels, achieving 85.76\% voxel saving), over 85\% of embedded voxels are eliminated. Even for datasets with high non-zero ratios, such as MSD Prostate (99.50\% nonzero pixels, 56.38\% voxel saving) and ACDC (97.72\% nonzero pixels, 59.11\% voxel saving), meaningful savings are still achieved. These results suggest that redundant background computation is pervasive across 3D medical imaging regardless of modality or anatomy, and that the proposed voxel-level pruning can recover a substantial portion of this wasted computation.

\begin{figure*}
	\centering  
	\includegraphics[width=\textwidth]{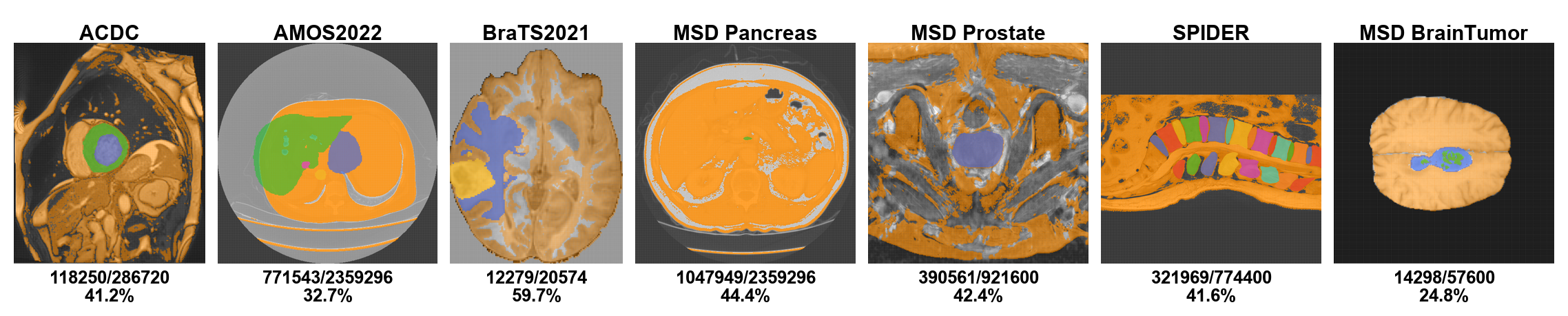}
    \caption{Nonvoid Voxelization visualization. First five columns: dataset-specific trained embeddings. Last two columns: zero-shot transfer (ACDC $\rightarrow$ SPIDER, BraTS2021 $\rightarrow$ MSD BrainTumor). Nonvoid/total voxel counts and ratios are shown below each volume.}
	\label{Exp_Fig_NVE_visualization}
\end{figure*}

\textbf{Nonvoid Voxel Embedding Visualization.} To qualitatively verify that Nonvoid Voxelization correctly retains anatomically relevant regions while discarding uninformative background, we visualize the embedded voxels in Fig.~\ref{Exp_Fig_NVE_visualization}. The visualization covers two scenarios. First, for the five segmentation benchmarks used throughout this work (BraTS2021, ACDC, MSD Prostate, MSD Pancreas, and AMOS2022), we show the nonvoid voxels produced by dataset-specific embeddings trained with Eq.~\eqref{eq:total_loss}. In all five cases, the occupancy map closely follows the anatomical boundaries without removing any labeled target voxels, confirming that the embedding layer learns to preserve clinically relevant content while suppressing background. Second, we test cross-dataset transferability by directly applying embedding parameters trained on one dataset to a related but different dataset without any retraining: parameters from ACDC are applied to SPIDER, and parameters from BraTS2021 are applied to MSD BrainTumor. In both cases, the transferred embedding still produces quite accurate Nonvoid Voxelization without notably abandoning target voxels, indicating that it captures a general notion of foreground occupancy rather than overfitting to dataset-specific anatomy. This transferability suggests that Nonvoid Voxelization has the potential to serve as a reusable preprocessing module across diverse clinical applications.

\subsection{Segmentation Quality} \label{sec:seg_quality}
To comprehensively validate the segmentation effectiveness of our GCNV-Net, we conduct experiments on five representative and widely-used 3D medical segmentation benchmarks: BraTS2021, ACDC, MSD Prostate, MSD Pancreas, and AMOS2022. We compare our method against a diverse set of advance baselines and state-of-the-art approaches, including nnU-Net \citep{nnUnet}, MedNeXt \citep{mednext}, E2ENet~\citep{e2enet}, SegFormer3D \citep{SegFormer3D}, nnFormer \citep{nnformer}, SwinUNETR \citep{swinUnetr}, PaNSegNet \citep{pansgnet}, and U-Mamba \citep{umamba}. We adopt Dice and IoU as primary overlap metrics, and additionally report HD95 and NSD as boundary-sensitive metrics to assess boundary quality, providing a comprehensive quantitative assessment.

\begin{table*}
    \fontsize{5.7bp}{8bp}\selectfont 
    \centering
    \caption{Quantitative comparison on BraTS2021, ACDC, MSD Pancreas, MSD Prostate, and AMOS2022 datasets. Dice is given in \%. Efficiency related metrics are evaluated on input volumes of size $1\times128\times128\times128$ (RTX4090, batch=1). QEA denotes the normalized quality-efficiency polygon enclosed area. \checkmark indicates the statistical significance of GCNV-Net versus each methods under paired Wilcoxon signed-rank test ($p < 0.05$, two-sided, Holm-corrected). $\uparrow$/$\downarrow$ denotes the preferred direction.}
    \label{Exp_Table_SegmentationResults}
    \vspace{1mm}
    \begin{tabular}{ccccccccccccccccc}
    \toprule
    \multirow{2}{*}{\textbf{Method}}
      & \multicolumn{2}{c}{\textbf{BraTS2021}} 
      & \multicolumn{2}{c}{\textbf{ACDC}} 
      & \multicolumn{2}{c}{\textbf{MSD Prostate}} 
      & \multicolumn{2}{c}{\textbf{MSD Pancreas}} 
      & \multicolumn{2}{c}{\textbf{AMOS2022}} 
      & \multirow{2}{*}{\textbf{\makecell{FLOPs \\ (G) $\downarrow$}}}
      & \multirow{2}{*}{\textbf{\makecell{Params \\ (M) $\downarrow$}}} 
      & \multirow{2}{*}{\textbf{\makecell{Latency \\ (ms) $\downarrow$}}} 
      & \multirow{2}{*}{\textbf{QEA $\uparrow$}} 
      & \multirow{2}{*}{\textbf{Rank $\downarrow$}} 
      & \multirow{2}{*}{\textbf{\makecell{Stat. \\ Sign.}}} \\
    \cmidrule(lr){2-3}\cmidrule(lr){4-5}\cmidrule(lr){6-7}\cmidrule(lr){8-9}\cmidrule(lr){10-11}
      & \textbf{Dice $\uparrow$} & \textbf{HD95 $\downarrow$} 
      & \textbf{Dice $\uparrow$} & \textbf{HD95 $\downarrow$} 
      & \textbf{Dice $\uparrow$} & \textbf{HD95 $\downarrow$} 
      & \textbf{Dice $\uparrow$} & \textbf{HD95 $\downarrow$} 
      & \textbf{Dice $\uparrow$} & \textbf{HD95 $\downarrow$} \\
    \midrule
    nnU-Net-ResEncL   & 91.19 & 1.33 & 91.65 & 0.78 & 73.91 & 15.79 & 67.86 & 5.89 & 89.42 & 4.27 & 1432.0 &  102.3 &  44.6 & 0.873 & 8  & \checkmark \\
    nnU-Net-ResEncM   & 91.24 & 1.37 & 92.01 & 0.77 & 74.07 & 15.26 & 68.12 & 5.70 & 88.73 & 4.61 & 1432.0 &  102.3 &  44.6 & 1.107 & 5  & \checkmark \\
    nnU-Net-NoResEnc  & 91.23 & 1.39 & 91.58 & 0.78 & 73.32 & 17.82 & 66.54 & 5.99 & 88.65 & 4.71 &  951.0 &   31.2 &  33.7 & 1.660 & 3  & \checkmark \\
    MedNeXt-L         & 91.46 & 1.33 & 92.60 & 0.74 & 72.90 & 19.25 & 69.30 & 5.77 & 89.77 & 3.94 &  584.7 &   63.0 & 189.3 & 1.191 & 4  & \checkmark \\
    MedNeXt-M         & 90.39 & 1.60 & 91.73 & 0.78 & 73.64 & 16.73 & 69.74 & 5.71 & 89.07 & 4.39 &  254.6 &   17.6 & 103.1 & 1.743 & 2  & \checkmark \\
    PaNSegNet         & 85.66 & 2.86 & 90.57 & 0.83 & 70.77 & 26.51 & 64.04 & 6.01 & 84.11 & 7.66 &  953.0 &   44.2 &  33.4 & 0.439 & 13 & \checkmark \\
    E2ENet            & 87.59 & 2.36 & 90.95 & 0.81 & 71.23 & 24.95 & 66.13 & 5.97 & 84.31 & 7.58 &  482.7 &   14.9 &  64.9 & 0.853 & 10 & \checkmark \\
    U-Mamba-Enc       & 89.93 & 1.74 & 91.22 & 0.80 & 70.52 & 27.36 & 66.75 & 6.00 & 88.37 & 4.87 &  976.6 &   59.6 & 114.7 & 0.727 & 11 & \checkmark \\
    U-Mamba-Bot       & 90.48 & 1.58 & 91.79 & 0.78 & 71.93 & 22.57 & 67.66 & 5.86 & 89.09 & 4.45 &  976.6 &   59.6 & 121.9 & 0.938 & 7  & \checkmark \\
    SwinUNETR         & 90.70 & 1.49 & 92.02 & 0.77 & 72.52 & 20.54 & 68.70 & 5.73 & 86.20 & 6.26 &  774.8 &   62.0 &  27.3 & 0.859 & 9  & \checkmark \\
    SegFormer3D       & 87.78 & 2.30 & 90.38 & 0.84 & 70.53 & 27.34 & 63.73 & 6.15 & 81.39 & 9.97 & \textbf{15.3} & \textbf{4.5} & \textbf{5.0} & 1.060 & 6 & \checkmark \\
    nnFormer          & 89.79 & 1.75 & 92.45 & 0.75 & 72.36 & 21.08 & 65.41 & 6.00 & 81.52 & 9.66 &  420.6 &  150.3 &  72.3 & 0.554 & 12 & \checkmark \\
    GCNV-Net (ours)   & \textbf{92.06} & \textbf{1.22} & \textbf{92.95} & \textbf{0.73} & \textbf{74.72} & \textbf{13.05} & \textbf{69.80} & \textbf{5.61} & \textbf{90.13} & \textbf{3.71} &  260.1 &   62.5 &  52.7 & \textbf{2.106} & \textbf{1} \\
    
    \bottomrule
    \end{tabular}
\end{table*}

\textbf{BraTS2021 Dataset.} As shown in Table~\ref{Exp_Table_SegmentationResults} (and Supplementary~Material~Table~1), our GCNV-Net achieves a new state-of-the-art on BraTS2021, with an average Dice of 92.06\%, IoU of 86.78\%, HD95 of 1.22, and NSD of 0.97. It attains the highest Dice on both whole tumor (WT: 94.44\%) and tumor core (TC: 93.17\%), surpassing strong CNN-based methods (e.g., nnU-Net-ResEncL: 93.23\% WT, 92.11\% TC), Transformer-based methods (e.g., SwinUNETR: 93.18\% WT, 91.48\% TC), and Mamba-based methods (e.g., U-Mamba-Bot: 92.77\% WT, 90.14\% TC). Beyond overlap metrics, GCNV-Net also achieves the lowest HD95 (1.22 vs. 1.33 for the second lowest competitor) and highest NSD (0.97 vs. 0.96 for most competitors), indicating tighter boundary adherence. These improvements are particularly notable on the irregular and heterogeneous tumor substructures characteristic of this benchmark.

\textbf{ACDC Dataset.} On the ACDC dataset (Supplementary~Material~Table~2), which involves cardiac segmentation of the right ventricular cavity (RVC), myocardium (MYO), and left ventricular cavity (LVC), GCNV-Net again sets a new state-of-the-art an average Dice of 92.95\%, IoU of 86.98\%, HD95 of 0.73, and NSD of 0.99. Compared to the strongest nnU-Net variant (nnU-Net-ResEncM: RVC 90.79\%, MYO 90.06\%), GCNV-Net improves Dice by +1.41\% (RVC: 92.20\%) and +1.82\% (MYO: 91.88\%). The boundary-sensitive metrics are equally strong, in which GCNV-Net achieves the lowest average HD95 and the highest NSD, confirming that the predicted contours closely match ground-truth boundaries. These gains on the thin and highly variable myocardial wall are particularly encouraging, as boundary accuracy in this region is directly relevant to clinical volumetric measurements.

\textbf{MSD Prostate Dataset.} For prostate segmentation (Supplementary~Material~Table~3), GCNV-Net reaches an average Dice, IoU, HD95, and NSD of 74.72\%, 62.55\%, 13.05, and 0.86, respectively, outperforming all competing methods and renew the state-of-the-art with no doute. On the clinically critical transition zone (TZ), where structural boundaries are ambiguous, GCNV-Net achieves 83.64\% Dice and 73.82\% IoU, improving over the previous best (SwinUNETR: 82.17\% Dice, 72.10\% IoU) by +1.47\% and +1.72\%, respectively. The HD95 improvement is even more pronounced, where GCNV-Net achieves 8.67 on TZ compared to 10.54 for SwinUNETR (second best), a 17.7\% reduction in boundary distance error. These results demonstrate the method's ability to handle low-contrast anatomical boundaries where precise delineation is most challenging.

\textbf{MSD Pancreas Dataset.} On pancreas segmentation (Supplementary~Material~Table~4), GCNV-Net once again establishes a new state-of-the-art. For the pancreas organ itself, GCNV-Net attains 84.71\% Dice, surpassing MedNeXt-M (84.18\%, second best) and SwinUNETR (83.99\%, third best). For pancreatic tumor segmentation, which is particularly challenging due to irregular shapes and small lesion sizes, GCNV-Net achieves a second best 54.89\% Dice, 43.70\% IoU, 7.53 HD95, and 0.76 NSD, competitive with the best prior results (MedNeXt-M: 55.29\% Dice, 44.28\% IoU, 7.58 HD95, and 0.77 NSD). Notably, on average, GCNV-Net achieves a highest Dice of 69.80\%, IoU of 58.40\%, HD95 of 5.61, and NSD of 0.85, confirming that both overlap and boundary quality is well maintained even on these highly irregular structures.

\textbf{AMOS2022 Dataset.} The AMOS2022 dataset includes annotations for 15 abdominal organs. (As detailed in Supplementary~Material~Table~5,) GCNV-Net achieves an average Dice of 90.13\%, IoU of 82.03\%, HD95 of 3.71, and NSD of 0.85. It attains the highest average Dice, IoU, and HD95 among all compared methods, outperforming the previous best (MedNeXt-L: 89.77\% Dice, 81.44\% IoU) by +0.36\% and +0.59\%, respectively. GCNV-Net ranks first or second on the majority of the 15 organs, with particularly strong results on both large-sized organs (e.g., Liver and Stomach) and small-sized structures (e.g., Postcava and Bladder). The average HD95 of 3.71 is the lowest among all methods (vs. 3.94 for MedNeXt-L), and the NSD of 0.85 is the second highest, just below the NSD for nnU-Net-ResEncL (0.86), further supporting that GCNV-Net produces spatially precise segmentations across organs of widely varying size and shape.

\begin{figure*}[t]
    \centering  
    \includegraphics[width=\textwidth]{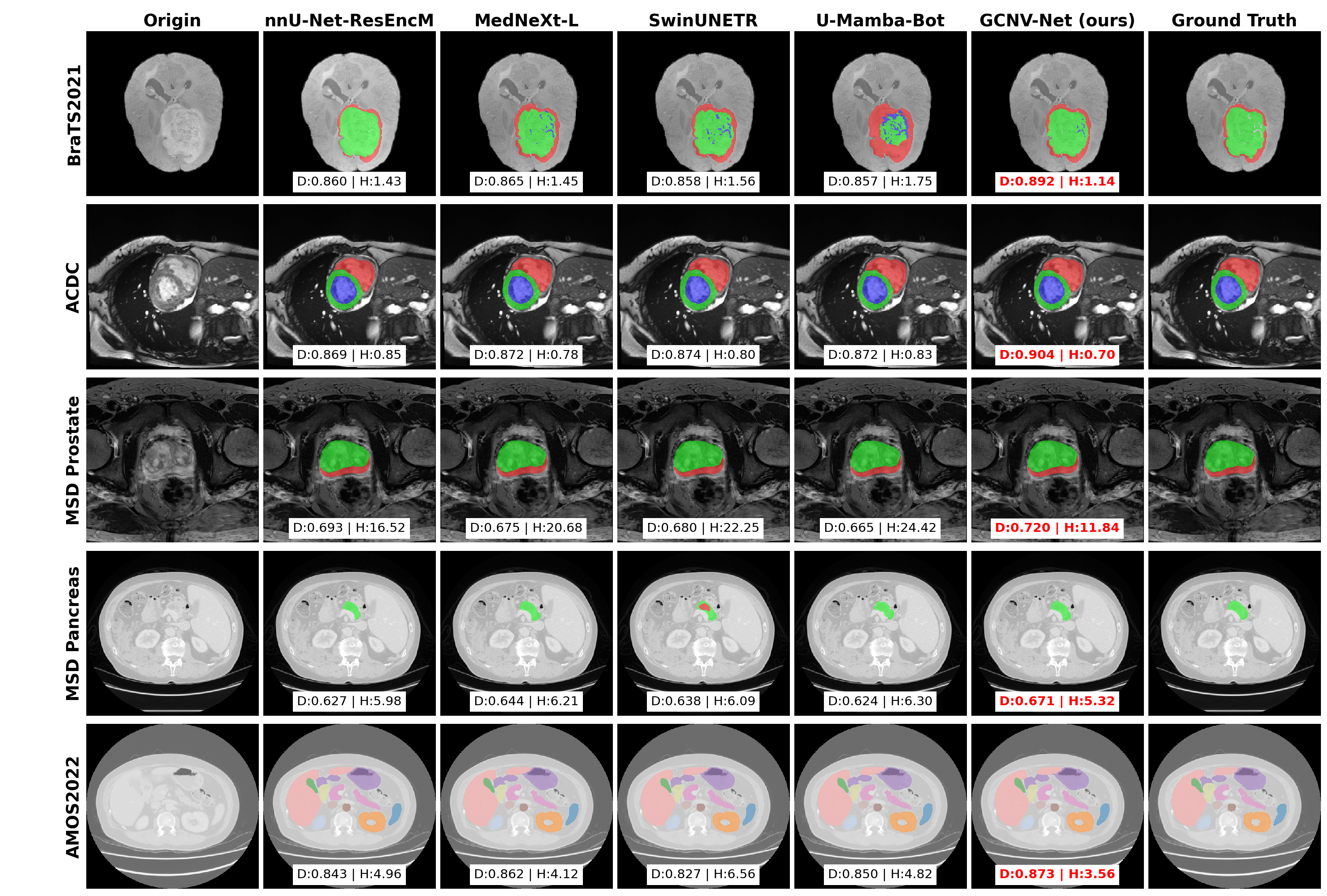}
    \caption{Qualitative segmentation comparison across all five benchmarks. Per-sample Dice (D) and HD95 (H) are shown below each prediction.}
    \label{Exp_Fig_Visualization}
\end{figure*}

\textbf{Visualization.} Fig. \ref{Exp_Fig_Visualization} provides representative visual comparisons across all five datasets. We can notice that GCNV-Net produces more precise boundaries and coherent segmentations compared to competing methods, particularly in tiny complex regions. For instance, on the BraTS2021 dataset, our method is the only one which can accurately identify the tiny unstructured region (presented in purple) in the scans. As can be seen, our method preserves detailed structure without sacrificing consistency, supporting further support the effectiveness of Geometrical Cross-Attention and dynamic voxel processing in preserving fine-grained boundary information.

\subsection{Quality Efficiency Trade-off} \label{sec:qua_eff_tradeoff}
While our primary focus is achieving top-tier segmentation accuracy, computational efficiency is equally important for clinical deployment. Compared to strong competitors with state-of-the-art accuracy such as nnU-Net (951.0G--1432.0G FLOPs) and SwinUNETR (774.8G FLOPs), GCNV-Net (260.1G FLOPs) achieves a 66\%-82\% reduction while simultaneously achieving higher Dice, IoU, NSD, and lower HD95 across all five benchmarks. This efficiency gain is largely attributable to Nonvoid Voxelization, which eliminates redundant computation on uninformative background regions before they enter the network.

\begin{figure}[t]
    \centering
    \includegraphics[width=3in]{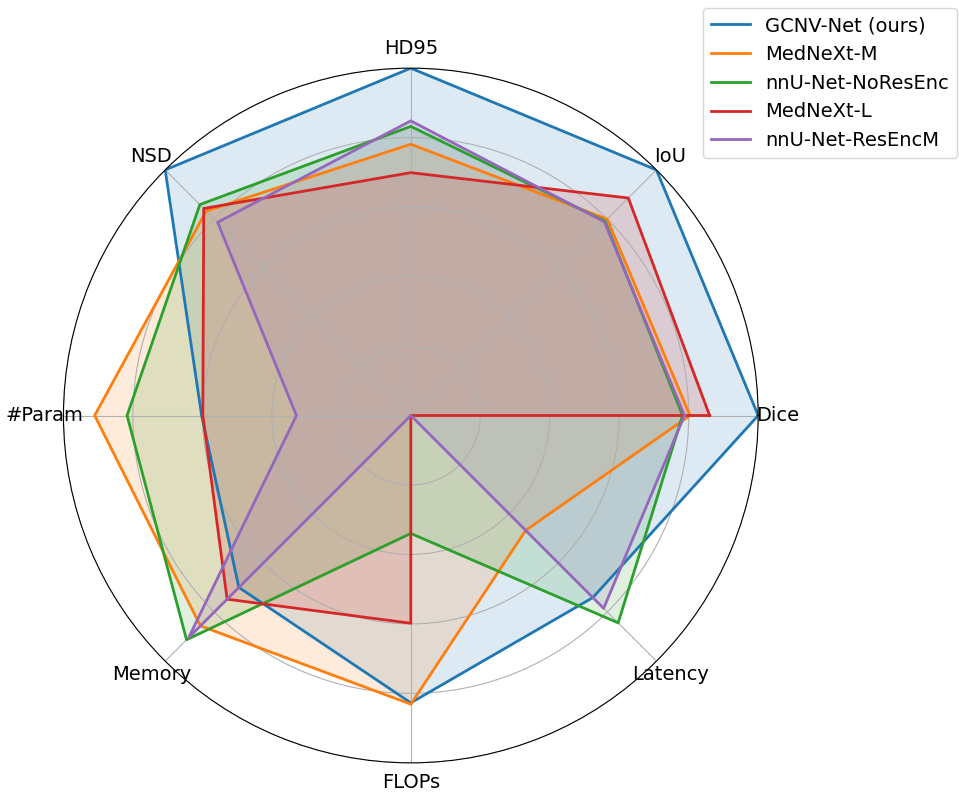}
    \caption{Quality-efficiency trade-off polygons (top-5 by enclosed area).}
    \label{Exp_TradeoffPolygon_Top5}
\end{figure}

However, evaluating accuracy and efficiency metrics separately does not adequately reflect the overall trade-off. To jointly quantify this balance, we construct a quality-efficiency polygon following a simple and commonly used approach. Specifically, we normalize all evaluation axes -- including Dice, IoU, HD95, NSD across datasets, as well as FLOPs, parameter count, memory consumption, and inference latency -- to the range $[0, 1]$. For metrics where lower is better (HD95, FLOPs, latency, etc.), we invert the normalized value so that a longer axis consistently indicates better performance. A polygon is then formed by connecting each method's normalized scores sequentially, and methods are ranked by the enclosed area: a larger area indicates a better overall accuracy-efficiency balance.

As shown in Fig.~\ref{Exp_TradeoffPolygon_Top5}, GCNV-Net exhibits the largest enclosed polygon area (QEA = 2.106) among all methods, substantially outperforming the second-ranked MedNeXt-M (QEA = 1.743) and third-ranked nnU-Net-NoResEnc (QEA = 1.660). The complete QEA scores for all methods are reported in Table~\ref{Exp_Table_SegmentationResults}. Notably, GCNV-Net dominates across both quality axes (Dice, IoU, HD95, NSD) and efficiency axes (FLOPs, latency), rather than excelling on only one side. This confirms that the proposed framework does not merely trade accuracy for speed or vice versa, but achieves a genuinely superior balance between the two — a property that is essential for practical clinical deployment.

\subsection{Statistical Significance}

\begin{figure}
	\centering  
	\includegraphics[width=3.3in]{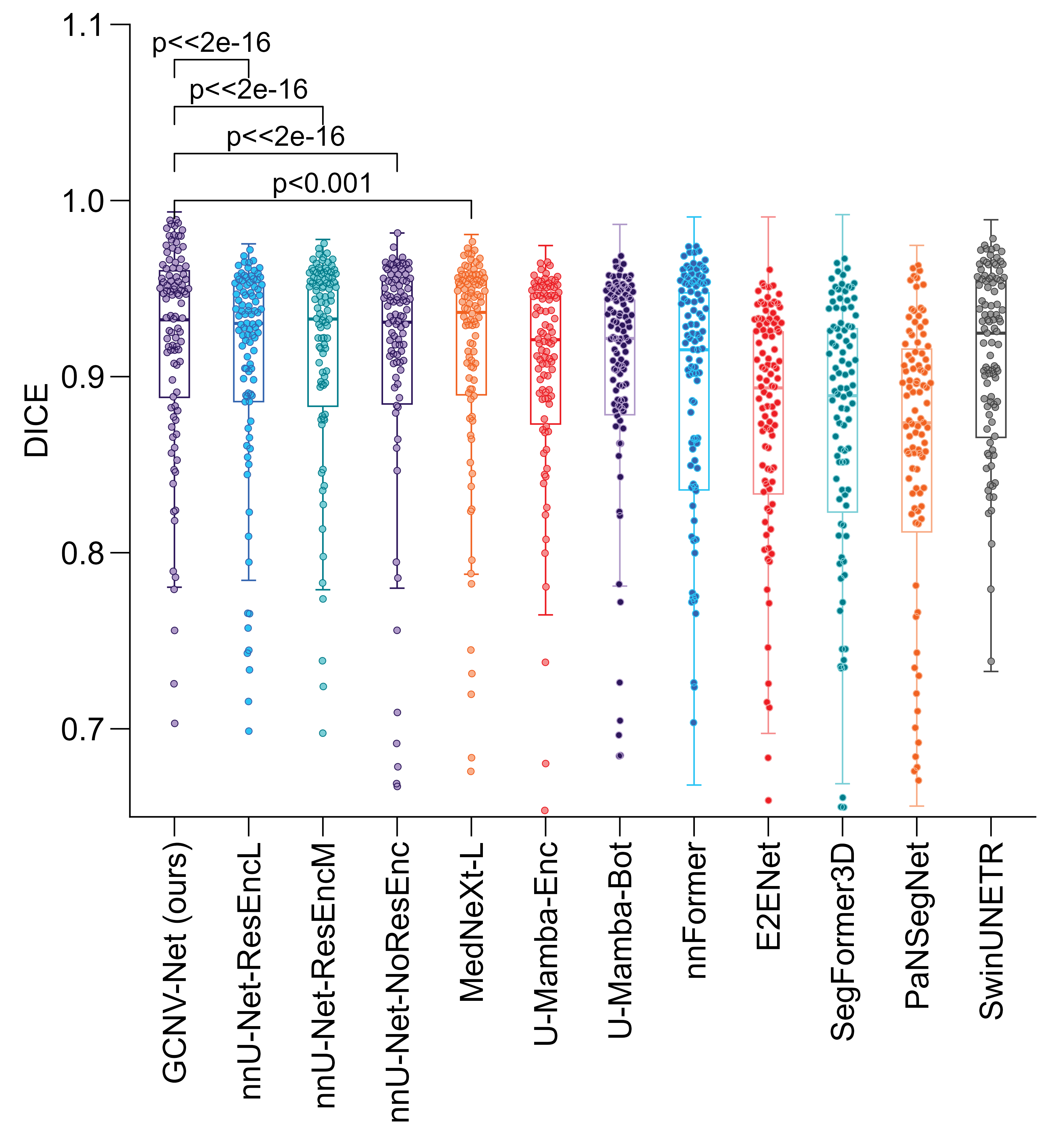}
	\caption{Per-case Dice distributions for different 3D segmentation methods. Brackets show representative p-values (top-4) from paired two-sided Wilcoxon signed-rank tests (Holm-corrected) comparing GCNV-Net with each competing methods. All comparisons yield p-values below 0.05, indicating statistically significant improvements.}
	\label{Exp_Fig_StasSign}
\end{figure}

To verify that the performance improvements achieved by GCNV-Net are statistically meaningful, we conducted paired two-sided Wilcoxon signed-rank tests on sample-wise Dice scores between GCNV-Net and each competing method. This non-parametric test is appropriate for paired comparisons on identical test samples and does not assume normality of segmentation metrics. To account for multiple hypothesis testing across different methods and datasets, Holm-Bonferroni correction was applied to control the family-wise error rate. A corrected significance level of $p < 0.05$ was used throughout. As summarized in Table~\ref{Exp_Table_SegmentationResults} and visualized in Fig.~\ref{Exp_Fig_StasSign} (as well as Supplementary~Material~Fig.~1), GCNV-Net demonstrates statistically significant improvements over all baseline and state-of-the-art methods, with all adjusted p-values remaining below the significance threshold.

\section{Ablation Study}
\begin{figure*}
    \centering
    \includegraphics[width=7in]{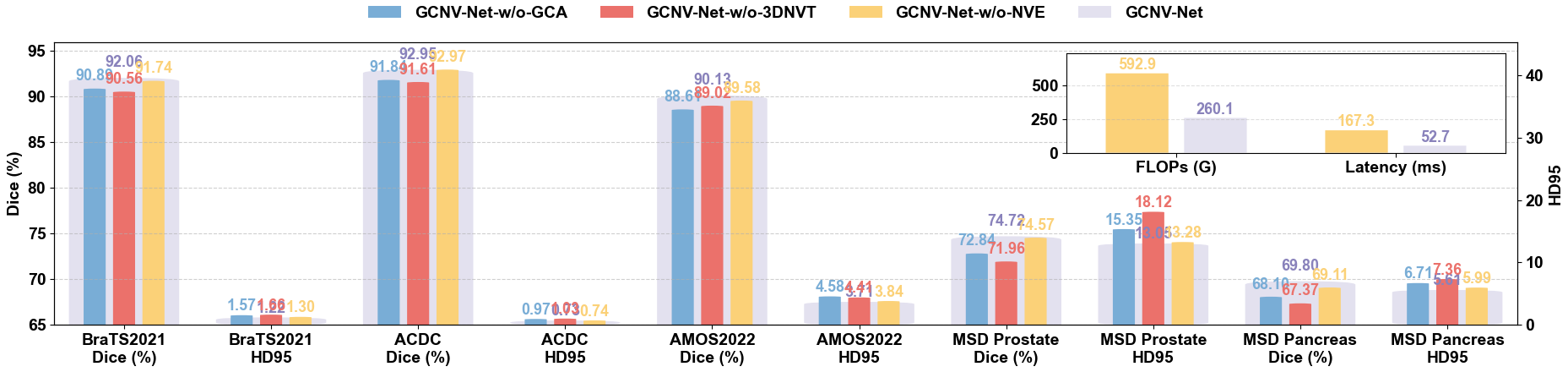}
    \caption{Mean Dice, HD95, FLOPs, and inference latency of the variants for ablation studies all benchmark datasets. FLOPs and latency are evaluated on input 3D volumes of size $1\times128\times128\times128$ (RTX4090, $batch=1$).}
    \label{Exp_Fig_AblationResults}
\end{figure*}

To isolate the contribution of each key component in GCNV-Net, we conduct ablation experiments by constructing three variants, each removing a single component:
\begin{itemize}
    \item GCNV-Net-w/o-GCA: This variant omits the Geometrical Cross-Attention upsampling and downsampling modules, relying solely on conventional max-pooling operations and MLPs for feature fusion across scales.
    \item GCNV-Net-w/o-3DNVT: In this variant, we replace the Tri-directional Dynamic Nonvoid Voxel Transformer blocks with traditional Swin Transformer blocks, eliminating dynamic partitioning along the three orthogonal axes.
    \item GCNV-Net-w/o-NVE: We remove the Nonvoid Voxelization step in this variant and use a widely used traditional voxelization method instead.
\end{itemize}

\subsection{Impact of Geometrical Cross-Attention} 
The segmentation performance (Dice and HD95) of these ablation variants on the BraTS2021, ACDC, AMOS022, MSD Prostate, and MSD Pancreas datasets is shown in Fig.~\ref{Exp_Fig_AblationResults}. As can be seen, on the BraTS2021 dataset, removing GCA significantly deteriorates segmentation performance, decreasing Dice and increasing HD95 from 92.06\% and 1.22 to 90.89\% and 1.57, respectively. The other two datasets follow the same trend. This notable degradation underscores the importance of explicitly integrating geometric positional information during multi-scale feature fusion. Our proposed Geometrical Cross-Attention module effectively enhances the network's ability to model spatial dependencies between preserved detailed anatomical structures, which conventional fusion methods alone fail to achieve.

\subsection{Impact of Tri-directional Dynamic Nonvoid Voxel Transformer} 
Similarly, omitting the 3DNVT module also leads to a noticeable decline in performance, with Dice and HD95 decreasing/increasing from 92.06 / 92.95 / 90.13 / 74.72 / 69.80 \% and 1.22 / 0.73 / 3.71 / 13.05 / 5.61 to 90.56 / 91.61 / 89.02 / 71.96 / 67.37 \% and 1.66 / 1.03 / 4.41 / 18.12 / 7.36, respectively. These results confirm that dynamically partitioning nonvoid voxels along multiple orthogonal axes significantly improves the ability of Transformer blocks to effectively capture complex spatial context. Without dynamic tri-directional partitioning, the model struggles to adequately handle densely structured medical voxels, thus reducing segmentation precision.

\subsection{Impact of Nonvoid Voxelization} 
Additionally, we analyze the computational efficiency improvements brought by Nonvoid Voxelization in terms of FLOPs and inference delay reduction. By comparing the FLOPs required for processing inputs embedded with conventional embedding (as used in \citep{swinUnetr, SegFormer3D}) and with our proposed Nonvoid Voxelization, we observed a significant reduction in computational cost by adopting the proposed strategy. Specifically, the use of Nonvoid Voxelization results in a significant FLOPs reduction of 56.13\% and a remarkable inference delay reduction of 68.49\%, clearly demonstrating its efficacy in reducing redundant computations associated with uninformative non-anatomical regions within the 3D medical image volume. Meanwhile, the accuracy difference between GCNV-Net and GCNV-Net-w/o-NVE is also negligible (within 0.02\% for Dice on the BraTS2021 dataset). Surprisingly, on all other datasets, the Nonvoid Voxelization even bring a small accuracy advance (up to 6.28\%) in both Dice and HD95, most likely due to the removal of noise through the abandon of non-anatomical regions. Together, proving that the efficiency improvement brought by our Nonvoid Voxelization is in fact a “free meal”.

\section{Conclusion}
In this paper, we proposed GCNV-Net, a novel and effective 3D medical image segmentation framework that synergistically integrates a Geometrical Cross-Attention mechanism, a Tri-directional Dynamic Nonvoid Voxel Transformer, and Nonvoid Voxelization. Extensive experiments conducted on five benchmark datasets -- BraTS2021, ACDC, AMOS022, MSD Prostate, and MSD Pancreas -- demonstrate that the proposed GCNV-Net achieves state-of-the-art segmentation performance across all benchmarks, consistently outperforming existing leading methods by substantial margins (up to 14.5\% on HD95) with statistical significance. Ablation studies confirm that each component contributes meaningfully: GCA improves Dice by up to 1.88\% through geometry-aware multi-scale fusion, 3DNVT improves Dice by up to 2.76\% through direction-specific spatial modeling, and Nonvoid Voxelization reduces FLOPs by 56.13\% and inference latency by 68.49\% while maintaining equivalent or even slightly improved accuracy (up to +0.69\% Dice and 6.28\% relative HD95 improvement).

\bibliographystyle{elsarticle-num}
\bibliography{reference}

\clearpage
\setcounter{section}{0}
\setcounter{subsection}{0}
\setcounter{subsubsection}{0}
\setcounter{figure}{0}
\setcounter{table}{0}

\section*{Supplementary Material}

\section{Segmentation Results}
\begin{figure*}
    \centering
    \begin{subfigure}{0.3\textwidth}
        \centering
        \includegraphics[width=\textwidth]{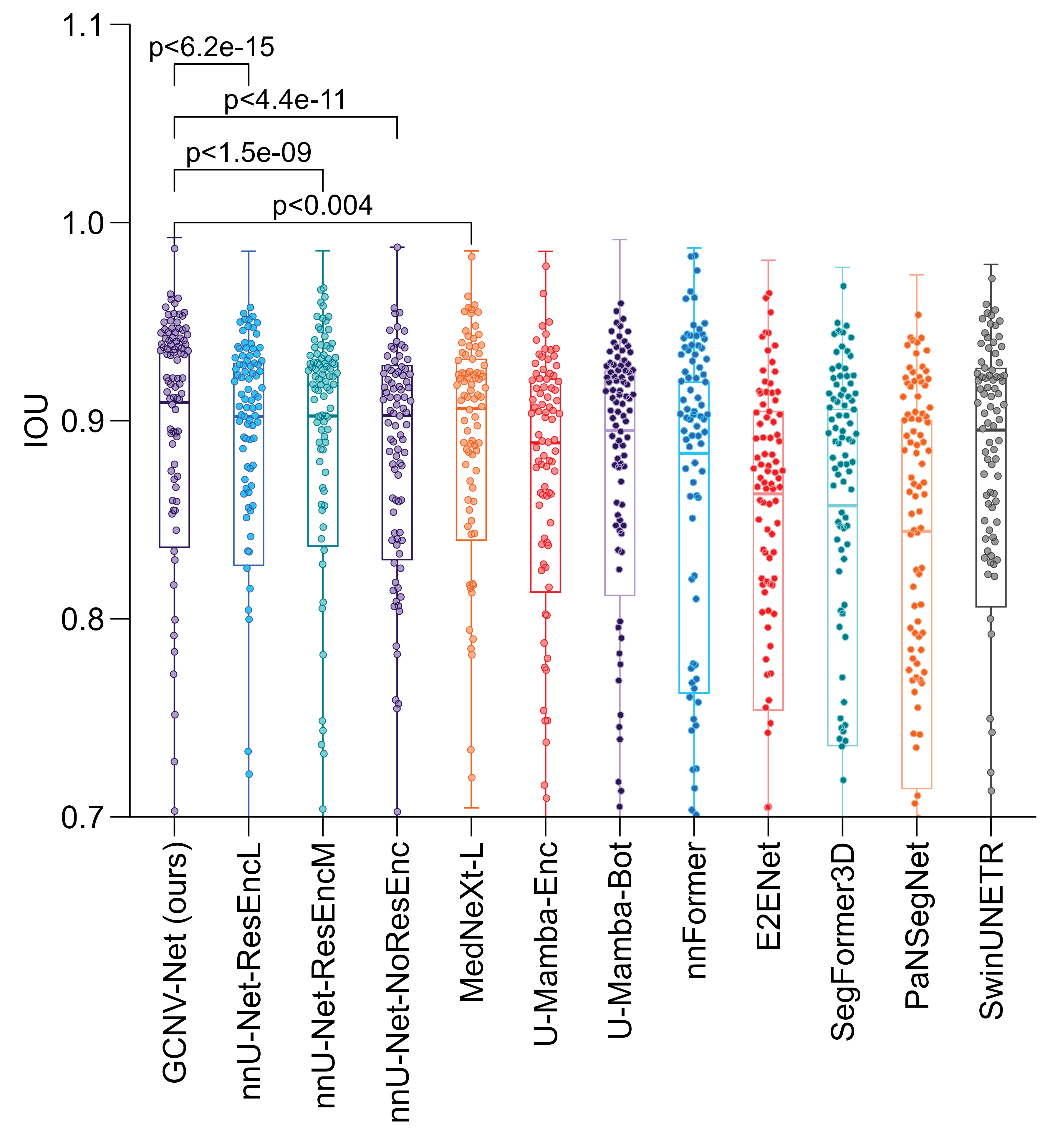}
        \caption{IoU}
    \end{subfigure}
    \hspace{1pt}
    \begin{subfigure}{0.3\textwidth}
        \centering
        \includegraphics[width=\textwidth]{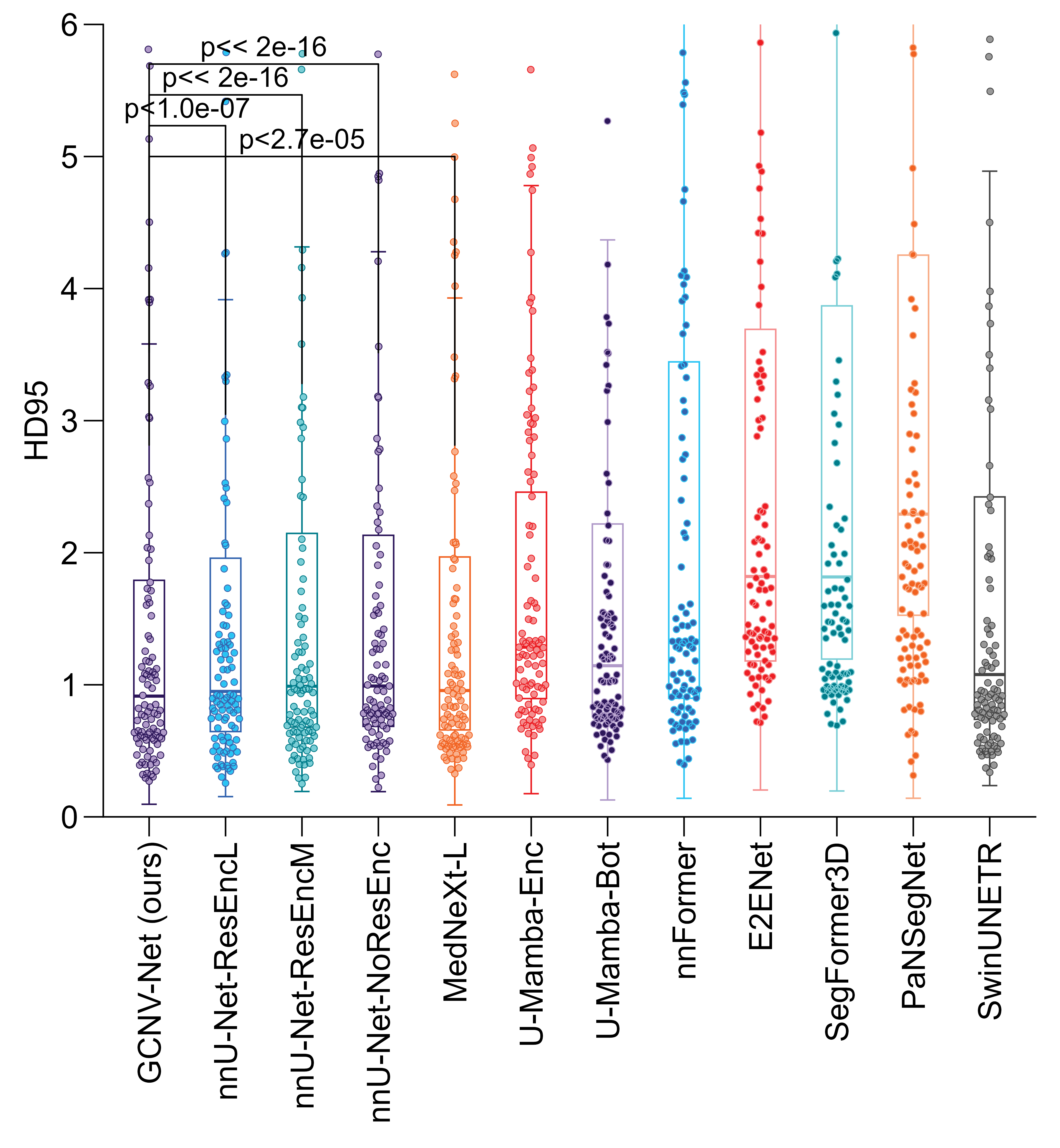}
        \caption{HD95}
    \end{subfigure}
    \hspace{1pt}
    \begin{subfigure}{0.3\textwidth}
        \centering
        \includegraphics[width=\textwidth]{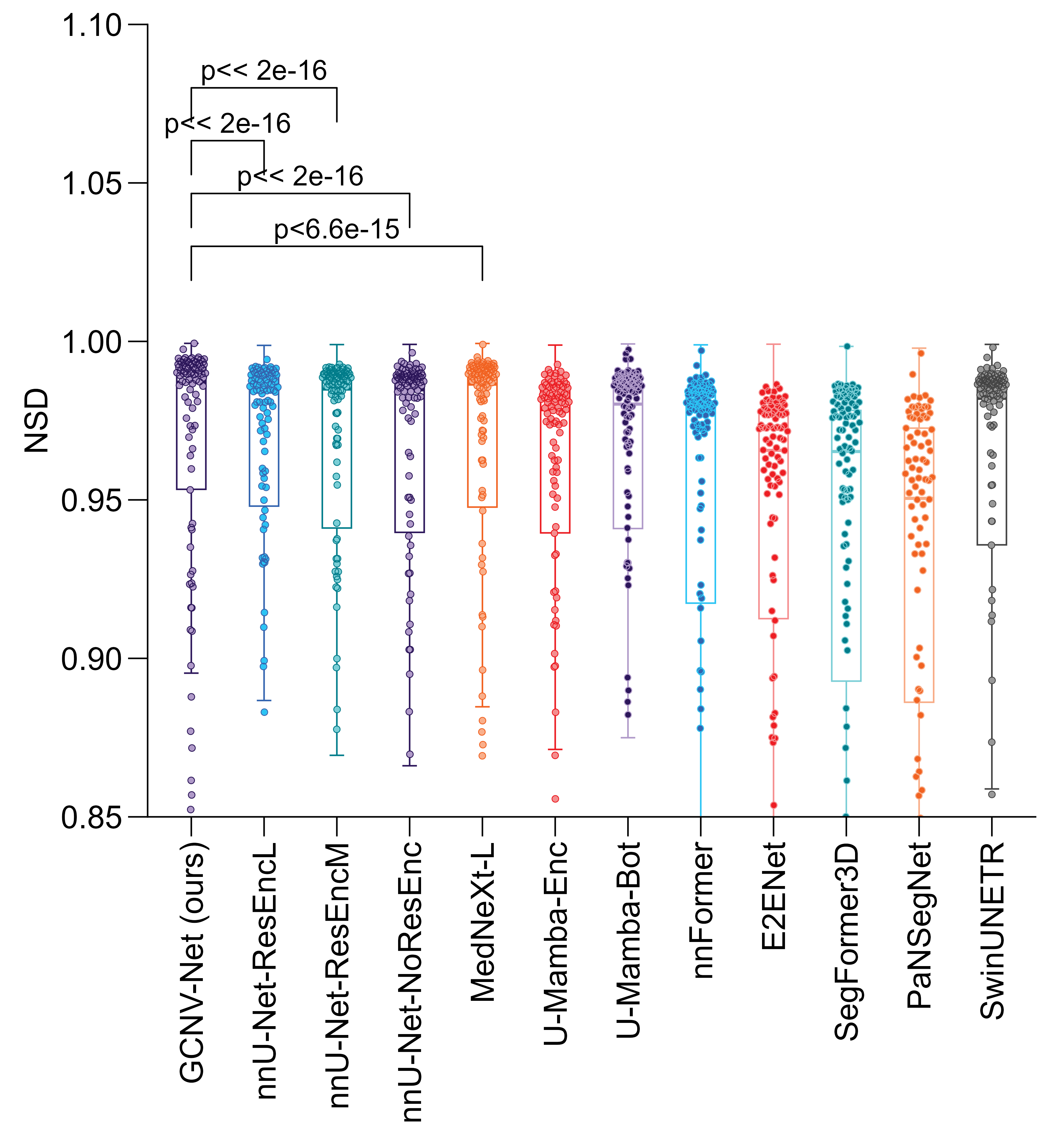}
        \caption{NSD}
    \end{subfigure}
    \caption{Per-case IoU, HD95, NSD distributions for different 3D segmentation methods. Brackets show representative p-values (top-4) from paired two-sided Wilcoxon signed-rank tests (Holm-corrected) comparing GCNV-Net with each competing methods. All comparisons yield p-values below 0.05, indicating statistically significant improvements.}
    \label{fig:StatisticalSignificance}
\end{figure*}

\begin{figure*}
    \centering
    \begin{subfigure}{\textwidth}
        \centering
        \includegraphics[width=\textwidth]{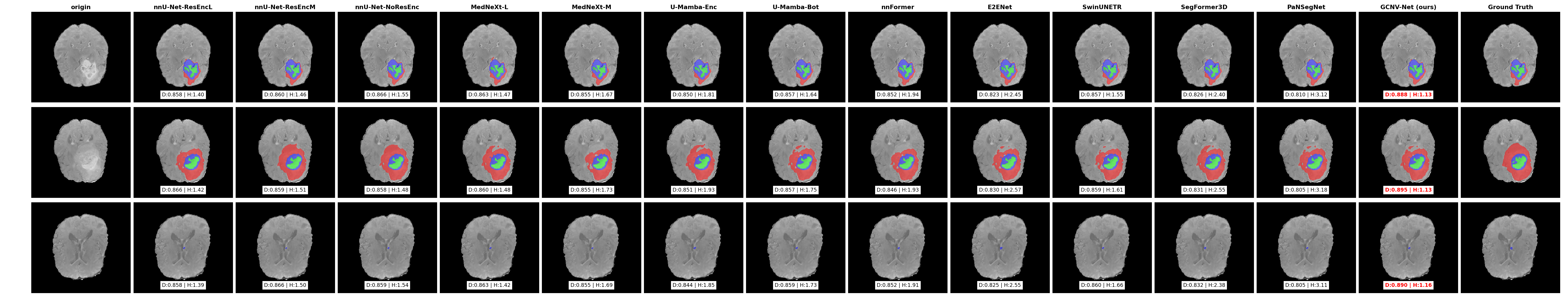}
        \caption{BraTS2021 Dataset}
    \end{subfigure}
    \vspace{0.5em}
    \begin{subfigure}{\textwidth}
        \centering
        \includegraphics[width=\textwidth]{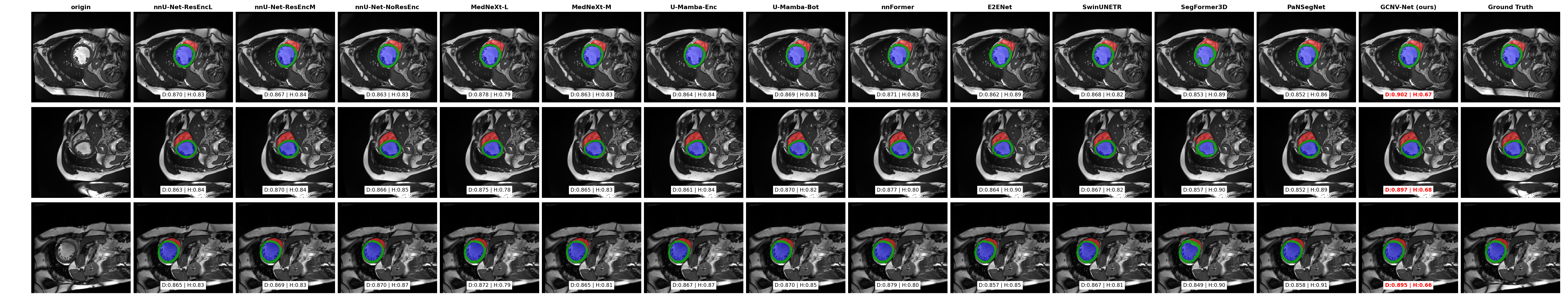}
        \caption{ACDC Dataset}
    \end{subfigure}
    \vspace{0.5em}
    \begin{subfigure}{\textwidth}
        \centering
        \includegraphics[width=\textwidth]{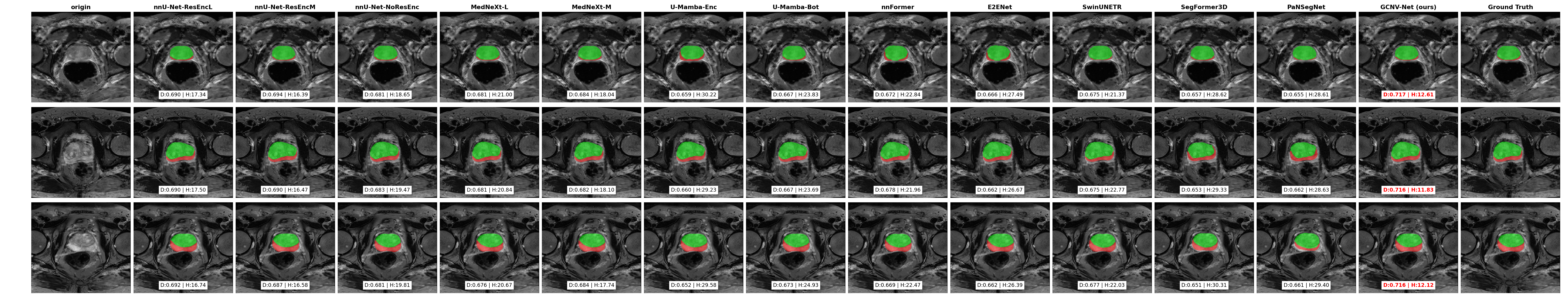}
        \caption{MSD Prostate Dataset}
    \end{subfigure}
    \vspace{0.5em}
    \begin{subfigure}{\textwidth}
        \centering
        \includegraphics[width=\textwidth]{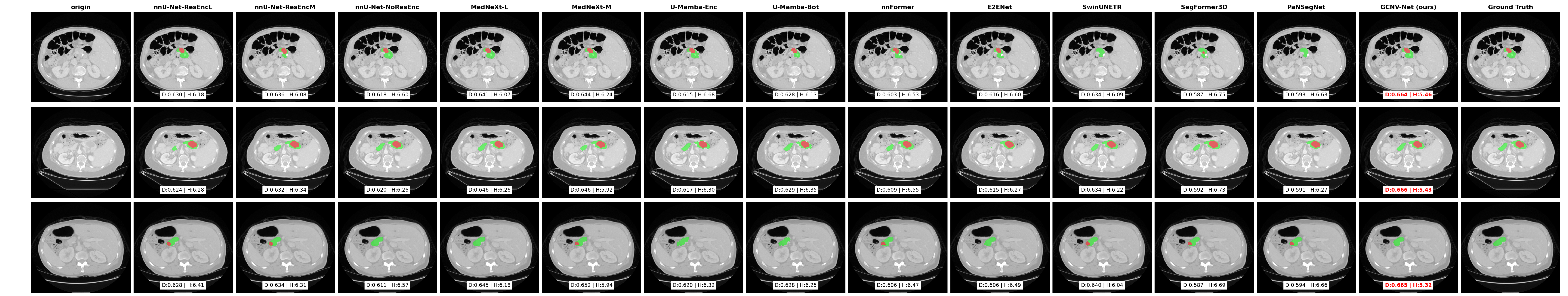}
        \caption{MSD Pancreas Dataset}
    \end{subfigure}
    \vspace{0.5em}
    \begin{subfigure}{\textwidth}
        \centering
        \includegraphics[width=\textwidth]{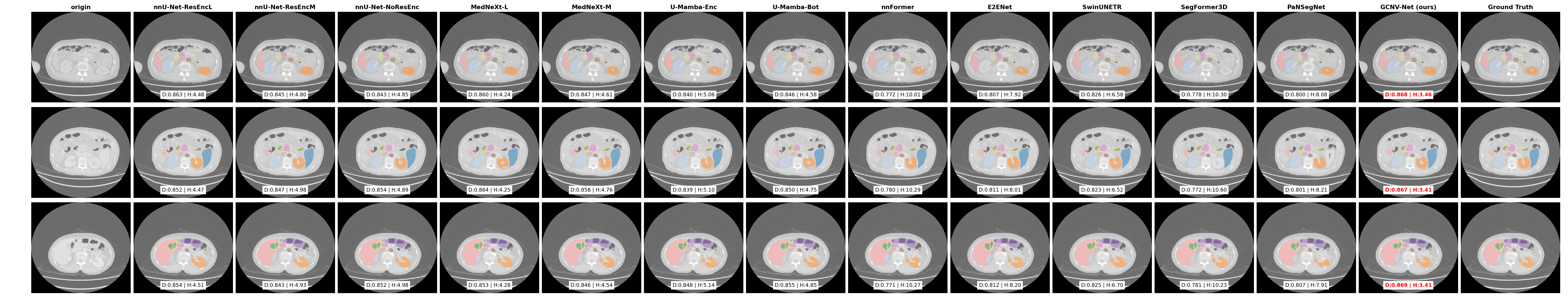}
        \caption{AMOS2022 Dataset}
    \end{subfigure}
    \caption{
        Visualized segmentation results on three representative samples from (a) BraTS2021, (b) ACDC, (c) MSD Prostate, (d) MSD Pancreas and (e) AMOS2022 datasets. Each subfigure includes predictions from twelve baseline models, our GCNV-Net, original input images (Origin), and ground truth. GCNV-Net demonstrates superior segmentation accuracy and boundary preservation across varied anatomical structures and imaging modalities.
    }
    \label{fig:combined_visualization}
\end{figure*}

\begin{figure}
    \centering
    \includegraphics[width=3.3in]{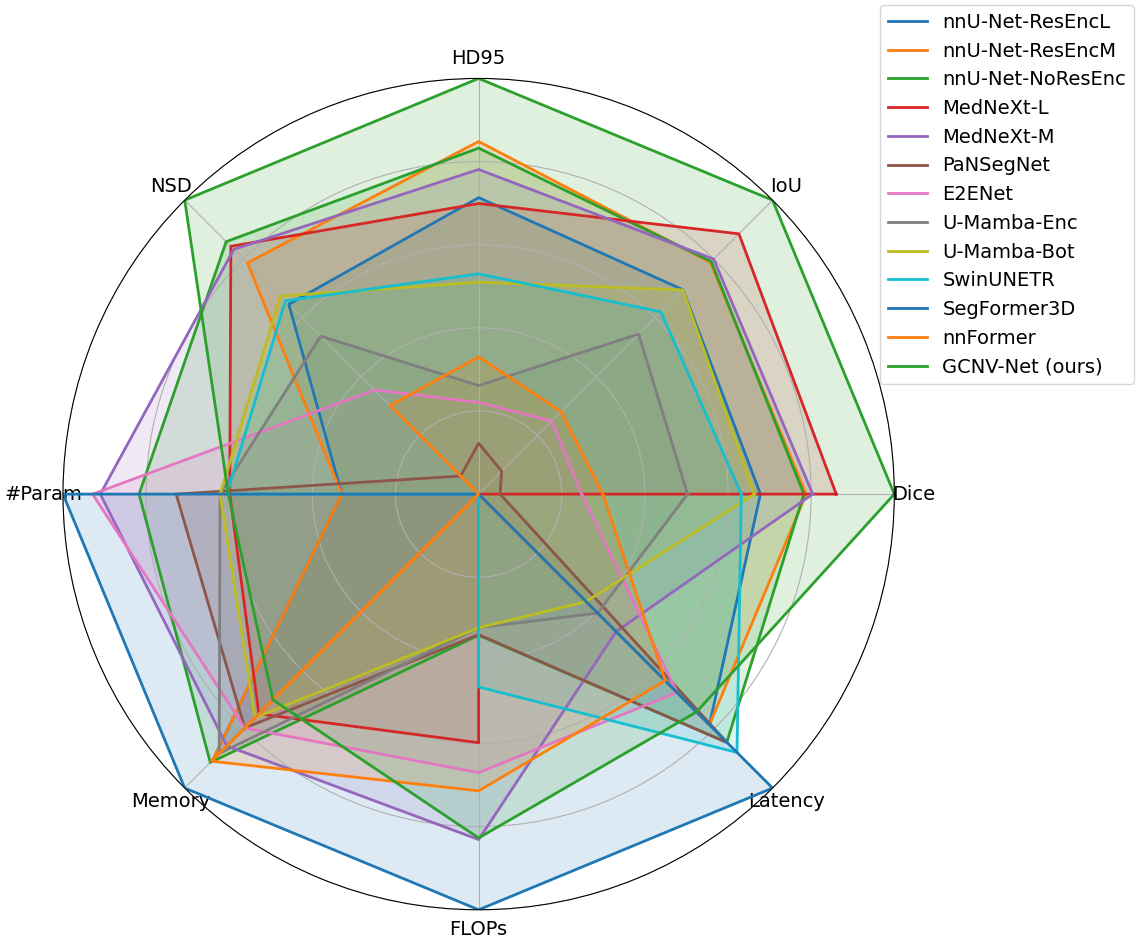}
    \caption{The quality efficiency trade off polygon.}
    \label{App_TradeoffPolygon}
\end{figure}

\begin{figure}
    \centering
    \includegraphics[width=3.3in]{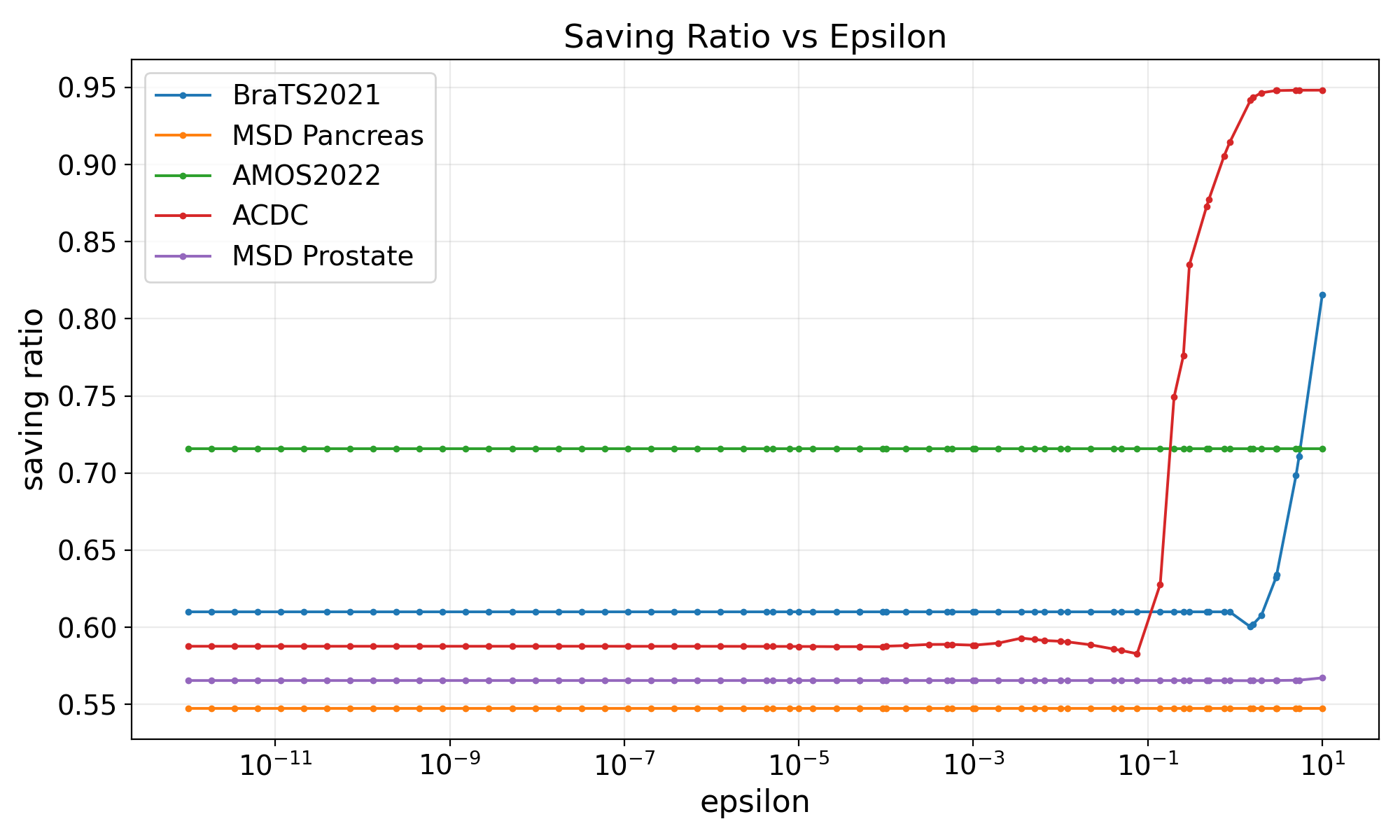}
    \caption{Embedded voxel saving of nonvoid voxelization versus under different $\epsilon$ (threshold hyperparameter).}
    \label{App_EpilonSensitivity}
\end{figure}

To complement the segmentation performance summaries reported in the main manuscript, this section presents a detailed per-class quantitative analysis across five benchmark datasets: BraTS2021, ACDC, MSD Prostate, MSD Pancreas, and AMOS2022. By reporting Dice, IoU, HD95, and NSD for individual anatomical structures, these results provide deeper insight into the fine-grained segmentation behavior of different methods and highlight the strengths of the proposed GCNV-Net in anatomically complex scenarios.

\begin{table*}[ht]
    \fontsize{6.1bp}{8bp}\selectfont 
    \centering
    \caption{Segmentation performance (Dice (\%), IoU (\%), HD95, and NSD) of different 3D segmentation methods on BraTS2021 datasets across three tumor substructures.}
    \label{Exp_Table_Accuracy_BraTS2021}
    \vspace{2pt}
    \begin{tabular}{ccccccccccccccccc}
        \toprule
        \multirow{2}{*}{\textbf{Method}} 
         & \multicolumn{4}{c}{\textbf{Whole Tumor}} 
         & \multicolumn{4}{c}{\textbf{Tumor Core}} 
         & \multicolumn{4}{c}{\textbf{Enhancing Tumor}} 
         & \multicolumn{4}{c}{\textbf{Average}} \\
        \cmidrule(lr){2-5} \cmidrule(lr){6-9} \cmidrule(lr){10-13} \cmidrule(lr){14-17}
         & \textbf{Dice $\uparrow$} & \textbf{IoU $\uparrow$} & \textbf{HD95 $\downarrow$} & \textbf{NSD $\uparrow$}
         & \textbf{Dice $\uparrow$} & \textbf{IoU $\uparrow$} & \textbf{HD95 $\downarrow$} & \textbf{NSD $\uparrow$}
         & \textbf{Dice $\uparrow$} & \textbf{IoU $\uparrow$} & \textbf{HD95 $\downarrow$} & \textbf{NSD $\uparrow$}
         & \textbf{Dice $\uparrow$} & \textbf{IoU $\uparrow$} & \textbf{HD95 $\downarrow$} & \textbf{NSD $\uparrow$} \\
        \midrule
        nnU-Net-ResEncL & 93.23 & 88.27 & 1.32 & 0.97 & 92.11 & 87.28 & 1.24 & 0.96 & 88.22 & 81.71 & 1.41 & 0.96 & 91.19 & 85.75 & 1.33 & 0.96 \\
        nnU-Net-ResEncM & 93.64 & 88.75 & 1.36 & 0.97 & 92.31 & 87.58 & 1.29 & 0.96 & 87.77 & 81.27 & 1.46 & 0.96 & 91.24 & 85.87 & 1.37 & 0.96 \\
        nnU-Net-NoResEnc & 93.87 & 89.00 & 1.32 & 0.97 & 92.31 & 87.57 & 1.30 & 0.96 & 87.51 & 80.91 & 1.56 & 0.96 & 91.23 & 85.83 & 1.39 & 0.96 \\
        MedNeXt-L & 93.19 & 88.38 & 1.36 & 0.98 & 91.67 & 86.81 & 1.19 & 0.97 & \textbf{89.52} & \textbf{83.26} & 1.45 & 0.96 & 91.46 & 86.15 & 1.33 & 0.97 \\
        MedNeXt-M & 92.71 & 87.85 & 1.49 & 0.97 & 90.03 & 84.88 & 1.41 & 0.96 & 88.44 & 82.08 & 1.89 & 0.94 & 90.39 & 84.94 & 1.60 & 0.96 \\
        U-Mamba-Enc & 92.29 & 87.35 & 1.71 & 0.96 & 89.54 & 84.21 & 1.45 & 0.96 & 87.96 & 81.51 & 2.06 & 0.94 & 89.93 & 84.35 & 1.74 & 0.95 \\
        U-Mamba-Bot & 92.77 & 87.87 & 1.52 & 0.97 & 90.14 & 84.96 & 1.29 & 0.96 & 88.53 & 82.04 & 1.93 & 0.95 & 90.48 & 84.96 & 1.58 & 0.96 \\
        nnFormer & 92.26 & 87.27 & 1.72 & 0.96 & 89.12 & 83.65 & 1.41 & 0.96 & 87.99 & 81.52 & 2.11 & 0.94 & 89.79 & 84.15 & 1.75 & 0.95 \\
        E2ENet & 89.65 & 84.33 & 2.41 & 0.95 & 87.07 & 80.97 & 1.61 & 0.94 & 86.05 & 79.31 & 3.05 & 0.92 & 87.59 & 81.54 & 2.36 & 0.94 \\
        SwinUNETR & 93.18 & 88.30 & 1.51 & 0.97 & 91.48 & 86.62 & 1.36 & 0.96 & 87.43 & 80.87 & 1.60 & 0.95 & 90.70 & 85.26 & 1.49 & 0.96 \\
        SegFormer3D & 89.75 & 84.47 & 2.33 & 0.95 & 87.64 & 81.57 & 1.60 & 0.94 & 85.96 & 79.37 & 2.96 & 0.93 & 87.78 & 81.80 & 2.30 & 0.94 \\
        PaNSegNet & 87.95 & 82.49 & 3.09 & 0.94 & 84.91 & 78.12 & 1.69 & 0.92 & 84.13 & 77.25 & 3.80 & 0.91 & 85.66 & 79.29 & 2.86 & 0.92 \\
        GCNV-Net (ours) & \textbf{94.44} & \textbf{89.64} & \textbf{1.21} & \textbf{0.98} & \textbf{93.17} & \textbf{88.57} & \textbf{1.15} & \textbf{0.97} & 88.55 & 82.14 & \textbf{1.29} & \textbf{0.97} & \textbf{92.06} & \textbf{86.78} & \textbf{1.22} & \textbf{0.97} \\
        \bottomrule
    \end{tabular}
\end{table*}

\textbf{BraTS2021 Dataset.}  
The BraTS2021 benchmark evaluates segmentation on three tumor subregions: Whole Tumor (WT), Tumor Core (TC), and Enhancing Tumor (ET). As detailed in Table~\ref{Exp_Table_Accuracy_BraTS2021}, GCNV-Net attains the highest scores across all four metrics for both WT (Dice 94.44\%, IoU 89.64\%, HD95 1.21, NSD 0.98) and TC (Dice 93.17\%, IoU 88.57\%, HD95 1.15, NSD 0.97), surpassing all CNN-based, Transformer-based, and Mamba-based competitors. For the ET subregion, MedNeXt-L achieves the highest Dice (89.52\%) and IoU (83.26\%); nevertheless, GCNV-Net obtains the best HD95 (1.29) and NSD (0.97) on ET, indicating tighter boundary adherence even where its overlap scores are not the highest. Overall, GCNV-Net achieves the best average performance across all four metrics (Dice 92.06\%, IoU 86.78\%, HD95 1.22, NSD 0.97).

\begin{table*}[ht]
    \fontsize{6.1bp}{8bp}\selectfont 
 	\centering
 	\caption{Segmentation performance (Dice (\%), IoU (\%), HD95, and NSD) of different 3D segmentation methods on ACDC dataset across three cardiac structures.}
 	\label{Exp_Table_Accuracy_ACDC}
    \vspace{2pt}
 	\begin{tabular}{ccccccccccccccccc}
 		\toprule
 		\multirow{2}{*}{\textbf{Method}} 
 		& \multicolumn{4}{c}{\textbf{Right Ventricular Cavity}} 
 		& \multicolumn{4}{c}{\textbf{Myocardium}} 
 		& \multicolumn{4}{c}{\textbf{Left Ventricular Cavity}} 
 		& \multicolumn{4}{c}{\textbf{Average}} \\
        \cmidrule(lr){2-5} \cmidrule(lr){6-9} \cmidrule(lr){10-13} \cmidrule(lr){14-17}
         & \textbf{Dice $\uparrow$} & \textbf{IoU $\uparrow$} & \textbf{HD95 $\downarrow$} & \textbf{NSD $\uparrow$}
         & \textbf{Dice $\uparrow$} & \textbf{IoU $\uparrow$} & \textbf{HD95 $\downarrow$} & \textbf{NSD $\uparrow$}
         & \textbf{Dice $\uparrow$} & \textbf{IoU $\uparrow$} & \textbf{HD95 $\downarrow$} & \textbf{NSD $\uparrow$}
         & \textbf{Dice $\uparrow$} & \textbf{IoU $\uparrow$} & \textbf{HD95 $\downarrow$} & \textbf{NSD $\uparrow$} \\
 		\midrule
 		nnU-Net-ResEncL & 90.21 & 82.85 & 0.94 & 0.98 & 89.75 & 81.63 & 1.02 & 0.99 & 94.98 & 90.64 & 0.38 & 0.99 & 91.65 & 85.04 & 0.78 & 0.99 \\
        nnU-Net-ResEncM & 90.79 & 83.72 & 0.93 & 0.98 & 90.06 & 82.10 & 1.00 & 0.99 & 95.17 & 90.92 & 0.37 & 1.00 & 92.01 & 85.58 & 0.77 & 0.99 \\
        nnU-Net-NoResEnc & 90.37 & 83.10 & 0.94 & 0.98 & 89.72 & 81.59 & 1.02 & 0.99 & 94.66 & 90.14 & 0.39 & 0.99 & 91.58 & 84.94 & 0.78 & 0.99 \\
        MedNeXt-L & 91.49 & 84.76 & 0.91 & 0.98 & 90.49 & 82.72 & 0.98 & 0.99 & \textbf{95.83} & \textbf{91.93} & \textbf{0.34} & \textbf{1.00} & 92.60 & 86.47 & 0.74 & 0.99 \\
        MedNeXt-M & 90.89 & 83.90 & 0.92 & 0.98 & 89.85 & 81.79 & 1.01 & 0.99 & 94.45 & 89.80 & 0.40 & 0.99 & 91.73 & 85.17 & 0.78 & 0.99 \\
        U-Mamba-Enc & 89.78 & 82.21 & 0.95 & 0.97 & 89.47 & 81.24 & 1.04 & 0.99 & 94.41 & 89.77 & 0.41 & 0.99 & 91.22 & 84.41 & 0.80 & 0.98 \\
        U-Mamba-Bot & 90.42 & 83.16 & 0.94 & 0.98 & 90.06 & 82.10 & 1.00 & 0.99 & 94.91 & 90.51 & 0.38 & 0.99 & 91.79 & 85.26 & 0.78 & 0.99 \\
        nnFormer & 91.90 & 85.41 & 0.89 & 0.98 & 90.00 & 81.97 & 1.00 & 0.99 & 95.44 & 91.33 & 0.35 & 1.00 & 92.45 & 86.24 & 0.75 & 0.99 \\
        E2ENet & 89.77 & 82.21 & 0.95 & 0.97 & 88.72 & 80.10 & 1.08 & 0.98 & 94.36 & 89.70 & 0.40 & 0.99 & 90.95 & 84.00 & 0.81 & 0.98 \\
        SwinUNETR & \textbf{92.50} & \textbf{86.59} & \textbf{0.75} & 0.98 & 91.54 & 84.61 & \textbf{0.78} & 0.99 & 93.86 & 89.60 & 0.39 & 0.99 & 92.02 & 85.60 & 0.77 & 0.99 \\
        SegFormer3D & 89.36 & 81.61 & 0.96 & 0.97 & 87.88 & 78.83 & 1.13 & 0.98 & 93.90 & 89.01 & 0.42 & 0.99 & 90.38 & 83.15 & 0.84 & 0.98 \\
        PaNSegNet & 89.42 & 81.70 & 0.96 & 0.97 & 88.30 & 79.46 & 1.11 & 0.98 & 93.99 & 89.13 & 0.42 & 0.99 & 90.57 & 83.43 & 0.83 & 0.98 \\
        GCNV-Net (ours) & 92.20 & 85.86 & 0.89 & \textbf{0.98} & \textbf{91.88} & \textbf{84.86} & 0.89 & \textbf{0.99} & 94.75 & 90.21 & 0.39 & 1.00 & \textbf{92.95} & \textbf{86.98} & \textbf{0.73} & \textbf{0.99} \\
 		\bottomrule
 	\end{tabular}
\end{table*}
 
\textbf{ACDC Dataset.}  
The ACDC dataset evaluates cardiac segmentation of the Right Ventricular Cavity (RVC), Myocardium (MYO), and Left Ventricular Cavity (LVC). As reported in Table~\ref{Exp_Table_Accuracy_ACDC}, GCNV-Net achieves the highest Dice and IoU on MYO (91.88\% and 84.86\%), the most anatomically challenging structure due to its thin and highly variable boundaries. On RVC, GCNV-Net (92.20\% Dice, 85.86\% IoU) ranks second only to SwinUNETR (92.50\% Dice, 86.59\% IoU) in overlap metrics, but achieves a higher NSD (0.98). For LVC, MedNeXt-L leads with 95.83\% Dice; GCNV-Net remains competitive at 94.75\%. On average, GCNV-Net achieves the best Dice (92.95\%), IoU (86.98\%), HD95 (0.73), and NSD (0.99) among all compared methods.

\begin{table*}[ht]
    \fontsize{7bp}{8bp}\selectfont 
    \centering
    \caption{Segmentation performance (Dice (\%), IoU (\%), HD95, and NSD) of different 3D segmentation methods on the MSD Prostate dataset across Peripheral Zone, Transition Zone, and their average.}
    \label{Exp_Table_Accuracy_Prostate}
    \vspace{2pt}
    \begin{tabular}{ccccccccccccc}
        \toprule
        \multirow{2}{*}{\textbf{Method}}
          & \multicolumn{4}{c}{\textbf{Peripheral Zone}}
          & \multicolumn{4}{c}{\textbf{Transition Zone}}
          & \multicolumn{4}{c}{\textbf{Average}} \\
        \cmidrule(lr){2-5} \cmidrule(lr){6-9} \cmidrule(lr){10-13}
         & \textbf{Dice $\uparrow$} & \textbf{IoU $\uparrow$} & \textbf{HD95 $\downarrow$} & \textbf{NSD $\uparrow$}
         & \textbf{Dice $\uparrow$} & \textbf{IoU $\uparrow$} & \textbf{HD95 $\downarrow$} & \textbf{NSD $\uparrow$}
         & \textbf{Dice $\uparrow$} & \textbf{IoU $\uparrow$} & \textbf{HD95 $\downarrow$} & \textbf{NSD $\uparrow$} \\
        \midrule
        nnU-Net-ResEncL & 66.11 & 51.63 & 16.97 & 0.81 & 81.72 & 72.02 & 14.62 & 0.90 & 73.91 & 61.83 & 15.79 & 0.85 \\
        nnU-Net-ResEncM & \textbf{66.35} & \textbf{51.81} & \textbf{15.91} & \textbf{0.81} & 81.79 & 72.12 & 14.61 & 0.90 & 74.07 & 61.97 & 15.26 & 0.85 \\
        nnU-Net-NoResEnc & 65.57 & 51.24 & 19.56 & 0.80 & 81.07 & 71.34 & 16.09 & 0.89 & 73.32 & 61.29 & 17.82 & 0.84 \\
        MedNeXt-L & 65.44 & 51.18 & 20.42 & 0.79 & 80.36 & 70.65 & 18.07 & 0.88 & 72.90 & 60.92 & 19.25 & 0.84 \\
        MedNeXt-M & 65.98 & 51.56 & 17.68 & 0.80 & 81.30 & 71.61 & 15.77 & 0.89 & 73.64 & 61.58 & 16.73 & 0.85 \\
        U-Mamba-Enc & 62.97 & 49.36 & 31.99 & 0.75 & 78.08 & 68.20 & 22.73 & 0.86 & 70.52 & 58.78 & 27.36 & 0.81 \\
        U-Mamba-Bot & 64.41 & 50.42 & 25.23 & 0.78 & 79.45 & 69.66 & 19.90 & 0.87 & 71.93 & 60.04 & 22.57 & 0.82 \\
        nnFormer & 64.78 & 50.68 & 23.44 & 0.78 & 79.95 & 70.19 & 18.73 & 0.88 & 72.36 & 60.43 & 21.08 & 0.83 \\
        E2ENet & 62.77 & 49.11 & 32.10 & 0.75 & 79.69 & 69.71 & 17.81 & 0.88 & 71.23 & 59.41 & 24.95 & 0.82 \\
        SwinUNETR & 62.88 & 49.05 & 30.54 & 0.76 & 82.17 & 72.10 & 10.54 & 0.91 & 72.52 & 60.58 & 20.54 & 0.83 \\
        SegFormer3D & 62.03 & 48.56 & 35.56 & 0.74 & 79.03 & 69.00 & 19.11 & 0.87 & 70.53 & 58.78 & 27.34 & 0.81 \\
        PaNSegNet & 62.16 & 48.65 & 34.85 & 0.74 & 79.39 & 69.36 & 18.17 & 0.88 & 70.77 & 59.00 & 26.51 & 0.81 \\
        GCNV-Net (ours) & 65.80 & 51.27 & 17.43 & 0.80 & \textbf{83.64} & \textbf{73.82} & \textbf{8.67} & \textbf{0.92} & \textbf{74.72} & \textbf{62.55} & \textbf{13.05} & \textbf{0.86} \\
        \bottomrule
    \end{tabular}
\end{table*}

\textbf{MSD Prostate Dataset.}  
Table~\ref{Exp_Table_Accuracy_Prostate} presents per-zone results for the Peripheral Zone (PZ) and Transition Zone (TZ). GCNV-Net demonstrates a clear advantage on the TZ, achieving the highest Dice (83.64\%), IoU (73.82\%), HD95 (8.67), and NSD (0.92), outperforming the second-best method (SwinUNETR: 82.17\% Dice, 72.10\% IoU, 10.54 HD95) by a substantial margin. On the PZ, nnU-Net-ResEncM leads with 66.35\% Dice; GCNV-Net is close at 65.80\%. The significant improvement on the clinically critical TZ, where boundaries are ambiguous and low-contrast, highlights the effectiveness of geometry-aware feature fusion. Overall, GCNV-Net attains the best average Dice (74.72\%), IoU (62.55\%), HD95 (13.05), and NSD (0.86).

\begin{table*}[t]
    \fontsize{7bp}{8bp}\selectfont 
    \centering
    \caption{Segmentation performance (Dice (\%), IoU (\%), HD95, and NSD) of different 3D segmentation methods on the MSD Pancreas dataset across Pancreas, Pancreatic Tumor, and their average.}
    \label{Exp_Table_Accuracy_Pancreas}
    \vspace{2pt}
    \begin{tabular}{ccccccccccccc}
        \toprule
        \multirow{2}{*}{\textbf{Method}}
          & \multicolumn{4}{c}{\textbf{Pancreas}}
          & \multicolumn{4}{c}{\textbf{Pancreatic Tumor}}
          & \multicolumn{4}{c}{\textbf{Average}} \\
        \cmidrule(lr){2-5} \cmidrule(lr){6-9} \cmidrule(lr){10-13}
         & \textbf{Dice $\uparrow$} & \textbf{IoU $\uparrow$} & \textbf{HD95 $\downarrow$} & \textbf{NSD $\uparrow$}
         & \textbf{Dice $\uparrow$} & \textbf{IoU $\uparrow$} & \textbf{HD95 $\downarrow$} & \textbf{NSD $\uparrow$}
         & \textbf{Dice $\uparrow$} & \textbf{IoU $\uparrow$} & \textbf{HD95 $\downarrow$} & \textbf{NSD $\uparrow$} \\
        \midrule
        nnU-Net-ResEncL & 82.85 & 71.23 & 3.34 & 0.91 & 52.86 & 42.00 & 8.43 & 0.73 & 67.86 & 56.62 & 5.89 & 0.82 \\
        nnU-Net-ResEncM & 83.01 & 71.33 & 3.92 & 0.91 & 53.23 & 42.38 & \textbf{7.48} & 0.73 & 68.12 & 56.85 & 5.70 & 0.82 \\
        nnU-Net-NoResEnc & 81.89 & 69.99 & 3.20 & 0.91 & 51.20 & 40.74 & 8.79 & 0.69 & 66.54 & 55.37 & 5.99 & 0.80 \\
        MedNeXt-L & 83.76 & 72.31 & 3.82 & 0.93 & 54.84 & 43.94 & 7.72 & 0.76 & 69.30 & 58.13 & 5.77 & 0.84 \\
        MedNeXt-M & 84.18 & 72.48 & 3.84 & 0.93 & \textbf{55.29} & \textbf{44.28} & 7.58 & \textbf{0.77} & 69.74 & 58.38 & 5.71 & 0.85 \\
        U-Mamba-Enc & 81.94 & 70.02 & 3.44 & 0.90 & 51.57 & 41.05 & 8.56 & 0.71 & 66.75 & 55.54 & 6.00 & 0.81 \\
        U-Mamba-Bot & 82.47 & 70.73 & 3.59 & 0.91 & 52.84 & 42.15 & 8.13 & 0.72 & 67.66 & 56.44 & 5.86 & 0.82 \\
        nnFormer & 80.02 & 67.84 & 2.76 & 0.89 & 50.80 & 40.62 & 9.24 & 0.68 & 65.41 & 54.23 & 6.00 & 0.78 \\
        E2ENet & 80.84 & 68.64 & 2.96 & 0.89 & 51.43 & 41.37 & 8.98 & 0.70 & 66.13 & 55.00 & 5.97 & 0.80 \\
        SwinUNETR & 83.99 & 72.51 & 3.63 & 0.93 & 53.40 & 42.44 & 7.82 & 0.74 & 68.70 & 57.47 & 5.73 & 0.83 \\
        SegFormer3D & 78.93 & 66.45 & 2.39 & 0.88 & 48.53 & 38.67 & 9.91 & 0.64 & 63.73 & 52.56 & 6.15 & 0.76 \\
        PaNSegNet & 79.44 & 67.31 & \textbf{2.25} & 0.88 & 48.63 & 38.91 & 9.77 & 0.65 & 64.04 & 53.11 & 6.01 & 0.77 \\
        GCNV-Net (ours) & \textbf{84.71} & \textbf{73.11} & 3.69 & \textbf{0.93} & 54.89 & 43.70 & 7.53 & 0.76 & \textbf{69.80} & \textbf{58.40} & \textbf{5.61} & \textbf{0.85} \\
        \bottomrule
    \end{tabular}
\end{table*}

\textbf{MSD Pancreas Dataset.}
Per-class results for the pancreas and pancreatic tumor are reported in Table~\ref{Exp_Table_Accuracy_Pancreas}. GCNV-Net achieves the highest pancreas Dice (84.71\%) and IoU (73.11\%), surpassing MedNeXt-M (84.18\%, 72.48\%). For the pancreatic tumor, MedNeXt-M leads with 55.29\% Dice and 44.28\% IoU; GCNV-Net is competitive at 54.89\% and 43.70\%, while achieving a lower HD95 (7.53 vs. 7.58). On average, GCNV-Net achieves the best Dice (69.80\%), IoU (58.40\%), HD95 (5.61), and NSD (0.85), confirming its effectiveness on organs with diverse anatomical characteristics and challenging pathological variability.

\begin{table*}[t]
    \fontsize{6.8bp}{8bp}\selectfont
    \centering
    \caption{Segmentation performance (Dice (\%), IoU (\%), HD95, and NSD) of different 3D segmentation methods on the AMOS2022 dataset across Spleen (Spl), Right Kidney (RK), Left Kidney (LK), Gall Bladder (GB), Esophagus (Eso), Liver (Liv), Stomach (Sto), Aorta (Aor), Postcava (IVC), Pancreas (Pan), Right Adrenal Gland (RAG), Left Adrenal Gland (LAG), Duodenum (Duo), Bladder (Bla), Prostate/Uterus (Pro/Ute), and their average (AVG).}
    \label{Exp_Table_Accuracy_AMOS}
    \begin{subtable}{\textwidth}
        \centering
        \caption{Dice (\%) $\uparrow$}
        \begin{tabular}{*{17}{c}}
            \toprule
            \textbf{Method} & \textbf{Spl} & \textbf{RK} & \textbf{LK} & \textbf{GB} & \textbf{Eso} & \textbf{Liv} &
            \textbf{Sto} & \textbf{Aor} & \textbf{IVC} & \textbf{Pan} & \textbf{RAG} & \textbf{LAG} &
            \textbf{Duo} & \textbf{Bla} & \textbf{Pro/Ute} & \textbf{AVG} \\
            \midrule
            nnU-Net-ResEncL & 95.01 & 89.17 & \textbf{82.57} & 83.50 & 84.34 & 91.98 & 88.55 & 93.81 & 92.87 & 83.77 & 85.15 & 95.49 & 89.93 & 94.89 & 90.20 & 89.42 \\
            nnU-Net-ResEncM & 94.48 & 88.45 & 81.67 & 82.79 & 83.56 & 91.14 & 87.79 & 93.11 & 92.25 & 82.98 & 84.30 & 95.13 & 89.25 & 94.34 & 89.69 & 88.73 \\
            nnU-Net-NoResEnc & 94.43 & 88.38 & 81.76 & 82.80 & 83.45 & 91.12 & 87.70 & 92.87 & 92.29 & 82.88 & 84.20 & 94.84 & 89.14 & 94.21 & 89.69 & 88.65 \\
            MedNeXt-L & 95.35 & 89.46 & 81.95 & 84.06 & 85.45 & 92.63 & 88.81 & 94.02 & 93.37 & 85.10 & 85.56 & 95.82 & 90.18 & 94.42 & 90.36 & 89.77 \\
            MedNeXt-M & 94.75 & 88.85 & 81.09 & 83.33 & 84.68 & 92.05 & 88.04 & 93.53 & 92.62 & 84.32 & 84.92 & 95.36 & 89.15 & 93.67 & 89.73 & 89.07 \\
            U-Mamba-Enc & 94.05 & 88.12 & 81.43 & 82.44 & 83.22 & 90.86 & 87.47 & 92.63 & 91.82 & 82.50 & 84.06 & 94.80 & 88.97 & 93.83 & 89.34 & 88.37 \\
            U-Mamba-Bot & 94.98 & 88.70 & 82.38 & 83.11 & 83.94 & 91.52 & 88.24 & 93.40 & 92.52 & 83.25 & 84.81 & 95.37 & 89.58 & 94.53 & 89.97 & 89.09 \\
            nnFormer & 87.18 & 81.07 & 74.62 & 75.68 & 76.40 & 84.44 & 80.50 & 85.73 & 84.99 & 75.96 & 77.01 & 87.64 & 81.94 & 86.98 & 82.61 & 81.52 \\
            E2ENet & 90.11 & 84.16 & 77.35 & 78.43 & 79.23 & 86.86 & 83.16 & 88.59 & 87.74 & 78.58 & 80.06 & 90.67 & 84.82 & 89.69 & 85.14 & 84.31 \\
            SwinUNETR & 91.63 & 86.01 & 79.58 & 80.54 & 80.78 & 88.79 & 85.37 & 90.17 & 89.44 & 80.70 & 81.75 & 92.71 & 86.72 & 91.63 & 87.14 & 86.20 \\
            SegFormer3D & 86.85 & 81.30 & 74.52 & 75.60 & 76.25 & 83.96 & 80.77 & 85.89 & 84.81 & 75.61 & 76.94 & 87.60 & 81.68 & 86.73 & 82.32 & 81.39 \\
            PaNSegNet & 89.71 & 83.99 & 77.14 & 78.30 & 78.99 & 86.70 & 83.14 & 88.53 & 87.64 & 78.55 & 79.65 & 90.18 & 84.40 & 89.65 & 85.07 & 84.11 \\
            GCNV-Net (ours) & \textbf{95.62} & \textbf{90.06} & 81.53 & \textbf{84.37} & \textbf{85.88} & \textbf{92.83} & \textbf{89.30} & \textbf{94.15} & \textbf{93.80} & \textbf{85.48} & \textbf{85.93} & \textbf{96.02} & \textbf{90.31} & \textbf{95.55} & \textbf{91.05} & \textbf{90.13} \\
            \bottomrule
        \end{tabular}
    \end{subtable}
    \hfill
    \begin{subtable}{\textwidth}
        \centering
        \caption{IoU (\%) $\uparrow$}
        \begin{tabular}{*{17}{c}}
            \toprule
            \textbf{Method} & \textbf{Spl} & \textbf{RK} & \textbf{LK} & \textbf{GB} & \textbf{Eso} & \textbf{Liv} &
            \textbf{Sto} & \textbf{Aor} & \textbf{IVC} & \textbf{Pan} & \textbf{RAG} & \textbf{LAG} &
            \textbf{Duo} & \textbf{Bla} & \textbf{Pro/Ute} & \textbf{AVG} \\
            \midrule
            nnU-Net-ResEncL & 90.20 & 80.17 & \textbf{70.03} & 71.38 & 72.64 & 84.87 & 79.17 & 88.06 & 86.40 & 71.79 & 73.85 & 91.08 & 81.42 & 89.99 & 81.86 & 80.86 \\
            nnU-Net-ResEncM & 89.24 & 78.99 & 68.71 & 70.33 & 71.47 & 83.42 & 77.94 & 86.81 & 85.32 & 70.61 & 72.57 & 90.41 & 80.29 & 88.98 & 81.01 & 79.74 \\
            nnU-Net-NoResEnc & 89.16 & 78.89 & 68.86 & 70.36 & 71.30 & 83.40 & 77.80 & 86.40 & 85.40 & 70.47 & 72.42 & 89.89 & 80.11 & 88.76 & 81.02 & 79.61 \\
            MedNeXt-L & 90.83 & 80.66 & 69.14 & 72.22 & 74.32 & 85.99 & 79.60 & 88.44 & 87.29 & 73.79 & 74.48 & 91.69 & 81.84 & 89.15 & 82.13 & 81.44 \\
            MedNeXt-M & 89.74 & 79.66 & 67.91 & 71.14 & 73.15 & 84.99 & 78.36 & 87.56 & 85.97 & 72.61 & 73.51 & 90.84 & 80.15 & 87.81 & 81.09 & 80.30 \\
            U-Mamba-Enc & 88.48 & 78.47 & 68.38 & 69.83 & 70.97 & 82.95 & 77.43 & 85.99 & 84.59 & 69.92 & 72.21 & 89.82 & 79.83 & 88.09 & 80.44 & 79.16 \\
            U-Mamba-Bot & 90.15 & 79.40 & 69.74 & 70.81 & 72.03 & 84.08 & 78.66 & 87.32 & 85.79 & 71.01 & 73.33 & 90.86 & 80.84 & 89.34 & 81.47 & 80.32 \\
            nnFormer & 77.04 & 67.93 & 59.27 & 60.64 & 61.58 & 72.83 & 67.13 & 74.79 & 73.66 & 60.99 & 62.38 & 77.76 & 69.17 & 76.72 & 70.14 & 68.80 \\
            E2ENet & 81.74 & 72.40 & 62.81 & 64.26 & 65.34 & 76.51 & 70.91 & 79.26 & 77.90 & 64.46 & 66.50 & 82.67 & 73.38 & 81.05 & 73.87 & 72.87 \\
            SwinUNETR & 84.29 & 75.19 & 65.83 & 67.16 & 67.49 & 79.57 & 74.20 & 81.84 & 80.63 & 67.39 & 68.88 & 86.15 & 76.29 & 84.29 & 76.95 & 75.74 \\
            SegFormer3D & 76.52 & 68.26 & 59.16 & 60.53 & 61.38 & 72.11 & 67.50 & 75.03 & 73.39 & 60.55 & 62.29 & 77.71 & 68.80 & 76.33 & 69.71 & 68.62 \\
            PaNSegNet & 81.09 & 72.15 & 62.53 & 64.09 & 65.02 & 76.27 & 70.90 & 79.16 & 77.74 & 64.42 & 65.93 & 81.87 & 72.75 & 80.98 & 73.77 & 72.58 \\
            GCNV-Net (ours) & \textbf{91.31} & \textbf{81.62} & 68.52 & \textbf{72.68} & \textbf{74.96} & \textbf{86.33} & \textbf{80.37} & \textbf{88.65} & \textbf{88.03} & \textbf{74.35} & \textbf{75.04} & \textbf{92.06} & \textbf{82.03} & \textbf{91.19} & \textbf{83.28} & \textbf{82.03} \\
            \bottomrule
        \end{tabular}
    \end{subtable}
    \hfill
    \begin{subtable}{\textwidth}
        \centering
        \caption{HD95 $\downarrow$}
        \begin{tabular}{*{17}{c}}
            \toprule
            \textbf{Method} & \textbf{Spl} & \textbf{RK} & \textbf{LK} & \textbf{GB} & \textbf{Eso} & \textbf{Liv} &
            \textbf{Sto} & \textbf{Aor} & \textbf{IVC} & \textbf{Pan} & \textbf{RAG} & \textbf{LAG} &
            \textbf{Duo} & \textbf{Bla} & \textbf{Pro/Ute} & \textbf{AVG} \\
            \midrule
            nnU-Net-ResEncL & 3.83 & 2.77 & 1.58 & 1.88 & 5.72 & 2.86 & 4.36 & 3.15 & 5.98 & 9.90 & 3.38 & 6.03 & 7.73 & 1.70 & 3.16 & 4.27 \\
            nnU-Net-ResEncM & 4.52 & 3.14 & 1.39 & 1.69 & 6.09 & 3.51 & 5.06 & 3.54 & 6.28 & 10.18 & 3.58 & 6.60 & 8.44 & 1.87 & 3.24 & 4.61 \\
            nnU-Net-NoResEnc & 4.62 & 3.13 & 1.17 & 1.63 & 6.50 & 3.32 & 4.91 & 3.45 & 6.78 & 10.56 & 3.61 & 7.62 & 8.40 & 1.81 & 3.17 & 4.71 \\
            MedNeXt-L & 3.58 & 2.60 & 1.60 & 1.96 & 5.38 & 2.79 & 4.23 & \textbf{2.88} & 5.03 & 8.23 & 3.50 & \textbf{4.83} & 7.39 & 2.10 & 2.93 & 3.94 \\
            MedNeXt-M & 4.03 & 2.64 & 1.71 & 1.79 & 5.33 & 2.72 & 4.90 & 3.34 & 6.82 & 10.32 & 3.52 & 5.82 & 7.36 & 2.48 & 3.06 & 4.39 \\
            U-Mamba-Enc & 5.77 & 3.54 & 1.44 & 1.94 & 6.17 & 3.54 & 5.26 & 4.23 & 6.95 & 10.23 & \textbf{3.28} & 7.30 & 8.01 & 2.21 & 3.22 & 4.87 \\
            U-Mamba-Bot & 3.84 & 2.76 & 1.44 & 1.88 & 6.40 & 3.19 & 4.86 & 3.55 & 6.39 & 10.17 & 3.54 & 6.17 & 7.53 & 1.93 & 3.12 & 4.45 \\
            nnFormer & 15.86 & 4.66 & \textbf{0.40} & \textbf{1.20} & 8.59 & 6.37 & 8.06 & 9.50 & 17.42 & 14.02 & 4.43 & 28.33 & 15.30 & 5.47 & 5.37 & 9.66 \\
            E2ENet & 9.71 & 3.80 & 1.20 & 1.66 & 8.28 & 5.54 & 7.38 & 7.65 & 12.69 & 13.89 & 3.68 & 17.35 & 12.30 & 4.10 & 4.53 & 7.58 \\
            SwinUNETR & 8.11 & 3.64 & 1.41 & 1.38 & 7.61 & 4.62 & 6.16 & 5.67 & 9.26 & 11.95 & 4.55 & 10.97 & 11.30 & 3.27 & 4.05 & 6.26 \\
            SegFormer3D & 16.99 & 5.00 & 0.99 & 1.61 & 8.56 & 7.15 & 7.57 & 9.03 & 18.60 & 16.26 & 4.54 & 25.10 & 16.75 & 5.83 & 5.51 & 9.97 \\
            PaNSegNet & 12.24 & 3.64 & 1.00 & 1.33 & 8.67 & 5.60 & 6.89 & 6.81 & 12.06 & 13.49 & 4.09 & 17.36 & 12.28 & 4.31 & 5.12 & 7.66 \\
            GCNV-Net (ours) & \textbf{3.24} & \textbf{2.56} & 1.74 & 1.60 & \textbf{4.57} & \textbf{2.56} & \textbf{4.03} & 2.89 & \textbf{4.94} & \textbf{8.11} & 3.33 & 4.97 & \textbf{7.28} & \textbf{1.29} & \textbf{2.64} & \textbf{3.71} \\
            \bottomrule
        \end{tabular}
    \end{subtable}
    \hfill
    \begin{subtable}{\textwidth}
        \centering
        \caption{NSD $\uparrow$}
        \begin{tabular}{*{17}{c}}
            \toprule
            \textbf{Method} & \textbf{Spl} & \textbf{RK} & \textbf{LK} & \textbf{GB} & \textbf{Eso} & \textbf{Liv} &
            \textbf{Sto} & \textbf{Aor} & \textbf{IVC} & \textbf{Pan} & \textbf{RAG} & \textbf{LAG} &
            \textbf{Duo} & \textbf{Bla} & \textbf{Pro/Ute} & \textbf{AVG} \\
            \midrule
            nnU-Net-ResEncL & 0.93 & 0.82 & 0.92 & 0.88 & \textbf{0.81} & 0.85 & 0.76 & 0.94 & 0.95 & \textbf{0.86} & 0.86 & 0.87 & 0.83 & 0.87 & 0.80 & \textbf{0.86} \\
            nnU-Net-ResEncM & 0.93 & 0.81 & \textbf{0.93} & 0.83 & 0.74 & 0.81 & 0.70 & 0.87 & \textbf{0.96} & 0.76 & 0.87 & 0.78 & 0.78 & 0.92 & 0.84 & 0.84 \\
            nnU-Net-NoResEnc & 0.86 & 0.82 & 0.88 & 0.87 & 0.71 & 0.87 & 0.71 & 0.85 & 0.87 & 0.80 & 0.83 & 0.85 & 0.80 & \textbf{0.98} & 0.79 & 0.83 \\
            MedNeXt-L & 0.85 & 0.80 & 0.87 & 0.91 & 0.76 & \textbf{0.90} & 0.71 & 0.83 & 0.91 & 0.82 & 0.89 & 0.88 & 0.81 & 0.97 & 0.77 & 0.84 \\
            MedNeXt-M & 0.91 & 0.78 & 0.88 & 0.95 & 0.77 & 0.83 & 0.69 & 0.81 & 0.88 & 0.81 & \textbf{0.91} & 0.86 & 0.83 & 0.89 & 0.77 & 0.84 \\
            U-Mamba-Enc & 0.92 & 0.73 & 0.84 & \textbf{0.99} & 0.70 & 0.79 & \textbf{0.80} & \textbf{1.00} & 0.84 & 0.78 & 0.84 & \textbf{0.93} & 0.76 & 0.90 & 0.85 & 0.84 \\
            U-Mamba-Bot & \textbf{0.99} & 0.86 & 0.92 & 0.86 & 0.71 & 0.88 & 0.74 & 0.90 & 0.82 & 0.83 & 0.85 & 0.80 & 0.77 & 0.93 & 0.82 & 0.85 \\
            nnFormer & 0.88 & 0.72 & 0.87 & 0.84 & 0.71 & 0.78 & 0.74 & 0.80 & 0.74 & 0.78 & 0.73 & 0.78 & 0.77 & 0.77 & 0.70 & 0.77 \\
            E2ENet & 0.88 & 0.81 & 0.81 & 0.86 & 0.71 & 0.80 & 0.74 & 0.88 & 0.93 & 0.76 & 0.77 & 0.82 & 0.78 & 0.89 & 0.75 & 0.81 \\
            SwinUNETR & 0.83 & 0.75 & 0.92 & 0.88 & 0.77 & 0.84 & 0.71 & 0.91 & 0.87 & 0.79 & 0.77 & 0.82 & 0.73 & 0.92 & 0.73 & 0.82 \\
            SegFormer3D & 0.85 & 0.76 & 0.83 & 0.71 & 0.68 & 0.76 & 0.67 & 0.87 & 0.73 & 0.69 & 0.76 & 0.71 & 0.77 & 0.84 & 0.77 & 0.76 \\
            PaNSegNet & 0.86 & 0.73 & 0.83 & 0.87 & 0.69 & 0.75 & 0.70 & 0.83 & 0.89 & 0.73 & 0.71 & 0.72 & 0.79 & 0.92 & 0.75 & 0.78 \\
            GCNV-Net (ours) & 0.96 & \textbf{0.86} & 0.89 & 0.95 & 0.78 & 0.86 & 0.78 & 0.82 & 0.87 & 0.77 & 0.76 & 0.84 & \textbf{0.85} & 0.93 & \textbf{0.86} & 0.85 \\
            \bottomrule
        \end{tabular}
    \end{subtable}
\end{table*}

\textbf{AMOS2022 Dataset.}
The AMOS2022 dataset includes annotations for 15 abdominal organs. As detailed in Table~\ref{Exp_Table_Accuracy_AMOS}, GCNV-Net achieves the highest average Dice (90.13\%), IoU (82.03\%), and HD95 (3.71), while its average NSD (0.85) ranks second to nnU-Net-ResEncL (0.86). GCNV-Net ranks first on 14 of 15 organs in Dice and IoU, with strong results across all organ sizes: large organs (e.g., Spleen 95.62\%, LAG 96.02\%), medium organs (e.g., Pancreas 85.48\%, Esophagus 85.88\%), and small structures (e.g., Postcava 93.80\%, Bladder 95.55\%). For the the one organ where GCNV-Net does not rank first in Dice — Left Kidney (81.53\% vs. 82.57\% for nnU-Net-ResEncL) — the margin is small. These results highlight the robustness of GCNV-Net across a wide range of anatomical structures.

\textbf{AMOS2022 Dataset.}
The AMOS2022 dataset includes annotations for 15 abdominal organs. As detailed in Table~\ref{Exp_Table_Accuracy_AMOS}, GCNV-Net achieves the highest average Dice (90.13\%), IoU (82.03\%), and HD95 (3.71), while its average NSD (0.85) ranks second to nnU-Net-ResEncL (0.86). GCNV-Net ranks first on 14 of 15 organs in Dice and IoU, with strong results across all organ sizes: large organs (e.g., Spleen 95.62\%, Liver 92.83\%), medium organs (e.g., Pancreas 85.48\%, Esophagus 85.88\%), and small structures (e.g., Left Adrenal Gland 96.02\%, Bladder 95.55\%). The only organ where GCNV-Net does not rank first in Dice is Left Kidney (81.53\% vs.\ 82.57\% for nnU-Net-ResEncL), a margin of just 1.04\%. These results highlight the robustness of GCNV-Net across a wide range of anatomical structures.

In summary, these per-class analyses demonstrate that GCNV-Net consistently achieves strong performance across complex and heterogeneous anatomical structures in diverse clinical domains. The combination of high overlap scores (Dice, IoU) with strong boundary metrics (HD95, NSD) confirms that the improvements are not limited to volumetric agreement but extend to contour precision.

\textbf{Statistical Significance.} In addition to the statistically significant improvements on Dice reported in the main manuscript, Fig.~\ref{fig:StatisticalSignificance} confirms that the improvements on IoU, HD95, and NSD are also statistically significant under the same testing protocol (paired two-sided Wilcoxon signed-rank test with Holm--Bonferroni correction, $p < 0.05$).

\section{Qualitative Visualization Results}

In addition to the representative examples shown in the main manuscript, Fig.~\ref{fig:combined_visualization} provides additional qualitative comparisons across all evaluated datasets.

On BraTS2021, GCNV-Net more accurately delineates irregular and heterogeneous tumor subregions, particularly for enhancing tumor areas, where competing methods often exhibit boundary blurring or under-segmentation.

For ACDC, the proposed method produces smoother and more anatomically consistent contours for the right ventricular cavity and myocardium, with visibly reduced leakage between adjacent cardiac structures, reflecting improved handling of thin and highly variable boundaries.

On MSD Prostate, GCNV-Net demonstrates clearer separation of the transition zone from surrounding tissues, preserving structural integrity in regions with low contrast and ambiguous boundaries, where baseline methods tend to over- or under-segment.

For MSD Pancreas, the model shows improved localization of both the pancreas and pancreatic tumors, effectively capturing small and irregular structures that are frequently missed or fragmented by competing approaches.

On AMOS2022 dataset, which involves multi-organ abdominal segmentation, GCNV-Net yields more coherent and spatially consistent segmentations across organs of varying sizes, highlighting its robustness in multi-scale, multi-class scenarios.

Overall, these qualitative results are consistent with the quantitative findings and further demonstrate the ability of GCNV-Net to generalize across diverse anatomical regions and imaging challenges.

\section{Radar Chart Details}

To jointly evaluate segmentation quality and computational efficiency, radar charts are constructed using normalized Dice, IoU, HD95, NSD, memory consumption, FLOPs, latency, and parameter size. For a given segmentation method $m$ with Dice score $Dice_m$ (the greater the better) and FLOPs $FLOPs_m$ (the smaller the better), we define their normalized line length $Dice_m'$ and $FLOPs_m'$ as:
\begin{equation}
    Dice_m' = \frac{Dice_m - Dice_{\min}}{Dice_{\max} - Dice_{\min}},
\end{equation}
\begin{equation}
    FLOPs_m' = \frac{FLOPs_{\max} - FLOPs_m}{FLOPs_{\max} - FLOPs_{\min}},
\end{equation}
where $Dice_{\max}$ and $Dice_{\min}$ denote the maximum and minimum Dice scores among all evaluated methods, and $FLOPs_{\max}$ and $FLOPs_{\min}$ denote the maximum and minimum FLOPs. The greater line length consistently indicate better performance along all axes. The complete radar chart with all methods is as Fig.~\ref{App_TradeoffPolygon}.

\section{Training Details}

All experiments are implemented within the nnU-Net v2 framework (v2.5.1 codebase), following its standard 3D full-resolution pipeline. Specifically, dataset fingerprinting, target spacing, resampling, and intensity normalization are all determined by nnU-Net's automated preprocessing. During training, patch-based online sampling and data augmentation---including random flipping, rotation, and intensity perturbation---are applied following nnU-Net defaults to improve model generalization. We do not redefine custom spacing rules, fixed global crop sizes, or a manually handcrafted preprocessing pipeline outside nnU-Net planning. Accordingly, batch size, patch size, and related training geometry are dataset-specific and automatically determined by \texttt{nnUNetPlans.json}, rather than being fixed across all datasets.

Our trainer extends the vanilla nnU-Net trainer with three additions: nonvoid voxelization--enabled network construction, runtime sparsity controls, and an auxiliary regularization term in the training loss. As described in Section~3.1 of the main manuscript, the training objective is:
\[
\mathcal{L}_{\mathrm{total}} = \mathcal{L}_{\mathrm{seg}} + \lambda_{\mathrm{nv}}\, r_{\mathrm{nv}},
\]
where $\mathcal{L}_{\mathrm{seg}}$ is the standard nnU-Net segmentation loss (Dice + Cross-Entropy with deep supervision), and $r_{\mathrm{nv}}$ is the soft nonvoid ratio computed via the sigmoid-based differentiable approximation of the hard occupancy map (see Section~3.1 of the main manuscript). The regularization weight is set to $\lambda_{\mathrm{nv}}=0.01$, the occupancy threshold to $\epsilon=10^{-5}$, and the sigmoid temperature to $\tau=1.0$ throughout all experiments. By minimizing $r_{\mathrm{nv}}$, the network receives continuous gradient pressure to suppress feature energy in uninformative regions while preserving strong activations in task-relevant areas, addressing the non-differentiability of hard occupancy gating. During inference, only the hard occupancy mask is applied, so the soft term introduces no additional cost at test time. Deep supervision is enabled in all main experiments, where auxiliary outputs at intermediate decoding stages are supervised with target tensors resized to match each resolution level.

\subsection{Modality-specific Background Constant}
\label{sec:bg_constant}
As noted in Section~3.1 of the main manuscript, the bias-free embedding convolution subtracts a per-channel background constant $b\in\mathbb{R}^{M}$ prior to convolution so that pure-background patches are mapped to strictly zero features. The derivation of $b$ depends on the normalization scheme employed by nnU-Net for each modality:

\begin{itemize}
\item \textbf{CT} (\texttt{CTNormalization}). All volumes share global intensity statistics (mean $\mu_{\mathrm{g}}$ and standard deviation $\sigma_{\mathrm{g}}$) from the dataset fingerprint. Air voxels, whose raw HU value equals the clip lower-bound $p_{0.5}$, are normalized to a fixed constant. We therefore set
\[
b_c = \frac{p_{0.5}^{(c)} - \mu_{\mathrm{g}}^{(c)}}{\sigma_{\mathrm{g}}^{(c)}},
\]
which is computed once from the fingerprint and stored as a fixed buffer in the network. This applies to AMOS2022 and MSD Pancreas.

\item \textbf{MRI with} \texttt{use\_mask\_for\_norm\,=\,True}. The z-score is computed using only foreground voxels, after which nnU-Net explicitly resets all background voxels to zero. Since background is already zero-valued, $b=0$ and the embedding convolution reduces to a plain bias-free convolution. This applies to BraTS2021.

\item \textbf{MRI with} \texttt{use\_mask\_for\_norm\,=\,False}. Each volume is independently z-scored with its own mean $\mu_i$ and standard deviation $\sigma_i$, shifting background voxels to $-\mu_i/\sigma_i$, a value that varies across cases. We estimate $b$ at runtime from the histogram mode of each preprocessed volume---the most frequent intensity value, which corresponds to background---at negligible computational cost. This applies to ACDC and MSD Prostate.
\end{itemize}

In all three cases, the key property is preserved: when a patch consists entirely of background voxels, the centered input $\mathcal{X}-b$ is identically zero, and the bias-free convolution produces a strictly zero output, enabling reliable nonvoid detection via the occupancy threshold $\epsilon$.

\subsection{Evaluation Protocol}
Cross-validation results are reported fold-wise under the standard nnU-Net 5-fold protocol. For final test-time inference, predictions from all five folds are aggregated following nnU-Net's default ensembling strategy. Test-time augmentation and connected component--based postprocessing are further applied to improve performance. The primary evaluation metric is Dice; boundary-sensitive metrics such as HD95 and NSD are additionally reported to support claims about contour and boundary quality. All experiments are conducted on a Linux server equipped with an NVIDIA RTX 4090 GPU.

\section{Analysis on Hyperparameter Sensitivity of Nonvoid Voxelization}
\label{App_Sec_Epsilon_Sensitivity}

The occupancy threshold $\epsilon$ in Eq.~(2) of the main manuscript determines whether an embedded voxel is classified as nonvoid or void. As discussed in Section~3.1, $\epsilon$ is designed to serve as a numerical guard against floating-point rounding rather than a tunable hyperparameter. To empirically verify this, we sweep $\epsilon$ across twelve orders of magnitude (from $10^{-11}$ to $10^{1}$) and measure the resulting embedded voxel saving ratio on all five segmentation benchmarks.

As shown in Fig.~\ref{App_EpilonSensitivity}, the saving ratio remains essentially constant across a wide range of $\epsilon$ values. For MSD Prostate, MSD Pancreas, and AMOS2022, the curves are nearly flat across the entire sweep from $10^{-11}$ to $10^{1}$, indicating that foreground and background feature norms are well separated on these datasets. For BraTS2021 and ACDC, the saving ratio is stable from $10^{-11}$ through $10^{-1}$, with a noticeable rise only beyond $10^{-1}$ as low-energy foreground voxels begin to be incorrectly classified as void. In both cases, $\epsilon = 10^{-5}$ — used throughout all our experiments — sits comfortably within the stable plateau, confirming that it does not require per-dataset tuning.

\end{document}